\documentclass{article}
\usepackage[preprint]{icml2026}

\usepackage{pifont}
\newcommand{\cmark}{\textcolor{green}{\ding{51}}}
\newcommand{\xmark}{\textcolor{red}{\ding{55}}}

\usepackage{color, colortbl}
\definecolor{Gray}{gray}{0.93}
\definecolor{Orange}{rgb}{1,0.5,0}
\definecolor{DGray}{gray}{0.83}
\definecolor{LightCyan}{rgb}{0.88,1,1}

\usepackage[T1]{fontenc}

\usepackage{microtype}
\usepackage{graphicx}
\usepackage{subfigure}
\usepackage{booktabs}

\usepackage{wrapfig}
\usepackage{pifont}

\usepackage{blindtext}
\usepackage{lipsum}

\usepackage{multirow}
\usepackage{graphicx}
\usepackage{listings}

\usepackage{bbm}

\usepackage [english]{babel}
\usepackage [autostyle, english = american]{csquotes}
\usepackage{amsmath}
\usepackage{amssymb}
\usepackage{mathtools}
\usepackage{amsthm}

\definecolor{mycitecolor}{HTML}{3498DC}
\definecolor{mylinkcolor}{HTML}{E74D3B}
\definecolor{myurlcolor}{HTML}{980000}
\definecolor{mydarkgreen}{HTML}{6a994e}
\definecolor{myorange}{HTML}{E7730D}
\definecolor{myblue}{HTML}{4594c1}
\usepackage[colorlinks=true,linkcolor=mylinkcolor,citecolor=mycitecolor,urlcolor=myurlcolor]{hyperref}

\usepackage[capitalize,noabbrev]{cleveref}
\usepackage{fontawesome}

\theoremstyle{plain}

\theoremstyle{definition}

\theoremstyle{remark}

\usepackage[textsize=tiny]{todonotes}

\usepackage{pifont}
\usepackage{url}
\usepackage[most]{tcolorbox}
\usepackage{lipsum}
\usepackage{wrapfig}
\usepackage{booktabs}
\usepackage{multirow,mathtools } 

\usepackage{adjustbox}
\MakeOuterQuote{"}

\usepackage{amsmath,amsfonts,bm}

\def\eqref#1{(\ref{#1})}

\def\1{\bm{1}}

\DeclareMathAlphabet{\mathsfit}{\encodingdefault}{\sfdefault}{m}{sl}
\SetMathAlphabet{\mathsfit}{bold}{\encodingdefault}{\sfdefault}{bx}{n}

\usepackage{color, colortbl}
\usepackage{xcolor}
\usepackage{colortbl}
\definecolor{mycitecolor}{HTML}{3498DC}
\definecolor{mylinkcolor}{HTML}{E74D3B}
\definecolor{myurlcolor}{HTML}{980000}
\definecolor{mydarkgreen}{HTML}{6a994e}
\definecolor{myorange}{HTML}{E7730D}
\definecolor{myblue}{HTML}{4594c1}
\definecolor{myyellow}{HTML}{FFDF00}
\definecolor{mypurple}{HTML}{C025CB}
\definecolor{RQframe}{HTML}{CFAB8D}
\definecolor{URAGColor}{HTML}{DE1A58}

\definecolor{sparse1}{HTML}{019E4F}
\definecolor{sparse2}{HTML}{E94127}
\definecolor{sparse3}{HTML}{017DC7}
\definecolor{dense1}{HTML}{3B6C73}
\definecolor{dense2}{HTML}{C1565E}
\definecolor{dense3}{HTML}{4682B4}

\definecolor{Appendixcolor}{HTML}{E74D3B}
\definecolor{Fgcolor}{HTML}{DD0303}
\definecolor{Sectioncolor}{HTML}{E74D3B}
\definecolor{Subsectioncolor}{HTML}{E74D3B}
\definecolor{Equationcolor}{HTML}{E74D3B}
\definecolor{Tablecolor}{HTML}{E74D3B}
\definecolor{Observationcolor}{HTML}{6a994e}
\definecolor{Algcolor}{HTML}{9305f2}
\definecolor{Promptcolor}{HTML}{FF6C0C}
\definecolor{RQcolor}{HTML}{001BB7}

\newcommand{\Appendix}{\textcolor{Appendixcolor}{Appendix}}
\newcommand{\Figure}{\textcolor{Fgcolor}{Figure}}
\newcommand{\Section}{\textcolor{Sectioncolor}{Section}}
\newcommand{\Sections}{\textcolor{Sectioncolor}{Sections}}

\newcommand{\Equation}{\textcolor{Equationcolor}{Equation}}
\newcommand{\Table}{\textcolor{Tablecolor}{Table}}
\newcommand{\Observation}{\textcolor{Observationcolor}{Observation}}
\newcommand{\Observations}{\textcolor{Observationcolor}{Observations}}
\newcommand{\Algorithm}{\textcolor{Algcolor}{Algorithm}}
\newcommand{\Prompt}{\textcolor{Promptcolor}{Prompt}}
\newcommand{\RQ}{\textcolor{RQcolor}{RQ}}

\usepackage[normalem]{ulem}
\usepackage{enumitem}
\usepackage{titletoc}

\newcommand{\Appendixref}[1]{\Appendix~{\hypersetup{linkcolor=Appendixcolor}\ref{#1}}}
\newcommand{\Figureref}[1]{\Figure~{\hypersetup{linkcolor=Fgcolor}\ref{#1}}}
\newcommand{\Sectionref}[1]{\Section~{\hypersetup{linkcolor=Sectioncolor}\ref{#1}}}

\newcommand{\Equationref}[1]{\Equation~{\hypersetup{linkcolor=Equationcolor}\ref{#1}}}
\newcommand{\Tableref}[1]{\Table~{\hypersetup{linkcolor=Tablecolor}\ref{#1}}}

\newcommand{\Algorithmref}[1]{\Algorithm~{\hypersetup{linkcolor=Algcolor}\ref{#1}}}
\newcommand{\Promptref}[1]{\Prompt~{\hypersetup{linkcolor=Promptcolor}\ref{#1}}}

\newcommand{\RQref}[1]{\RQ{\hypersetup{linkcolor=RQcolor}\ref{#1}}}

\usepackage[most]{tcolorbox}

\newcounter{rq}

\newcommand{\researchquestion}[1]{%
  \refstepcounter{rq}%
  \textcolor{RQcolor}{\textbf{RQ\therq:}} #1%
}

\newcounter{observation}
\renewcommand{\theobservation}{\arabic{observation}}

\newenvironment{observation}[1][]%
{%
    \refstepcounter{observation}
    \tcolorbox[
        enhanced,
        colback=white,
        colframe=white,
        leftrule=0.4mm,
        rightrule=0.4mm,
        toprule=0.4mm,
        bottomrule=0.4mm,
        arc=0mm,
        left=0pt,
        right=0pt,
        top=2pt,
        bottom=2pt,
        breakable,
        borderline north={0.4mm}{0pt}{mydarkgreen!80!black},
        borderline south={0.4mm}{0pt}{mydarkgreen!80!black}
    ]
    \textbf{\textcolor{mydarkgreen!80!black}{\textit{Observation~\theobservation}}}
    \ifx\relax#1\relax\else~(\textit{#1}).\fi%
}
{%
    \endtcolorbox
}

\newcounter{promptbox}

\newenvironment{promptbox}[1][]{
  \refstepcounter{promptbox}
  \begin{tcolorbox}[
    enhanced,
    colback=gray!5,
    colframe=black!70,
    title=\textbf{Prompt~\thepromptbox\ifx\relax#1\relax\else: #1\fi},
    fontupper=\small,
    arc=2mm,
    boxrule=1pt,
  ]
}{
  \end{tcolorbox}
}

\newcommand{\keyquestion}[1]{%
\begin{tcolorbox}[
    enhanced,
    colback=myblue!8!white,
    colframe=myblue,
    leftrule=2mm,
    rightrule=0mm,
    toprule=0mm,
    bottomrule=0mm,
    arc=0mm,
    left=5pt,
    right=5pt,
    top=5pt,
    bottom=5pt,
    breakable,
    leftlower=2mm,
    leftupper=2mm,
]
#1
\end{tcolorbox}
}

\newcommand{\contribution}[1]{%
\begin{tcolorbox}[
    enhanced,
    colback=mypurple!8!white,
    colframe=mypurple,
    leftrule=2mm,
    rightrule=0mm,
    toprule=0mm,
    bottomrule=0mm,
    arc=0mm,
    left=5pt,
    right=5pt,
    top=5pt,
    bottom=5pt,
    breakable,
    leftlower=2mm,
    leftupper=2mm
]
\normalsize 
\textit{#1}
\end{tcolorbox}
}

\usepackage{algorithm}
\usepackage{amsmath}
\usepackage{amssymb}
\usepackage{url}
\usepackage{array}
\usepackage{microtype}
\usepackage{graphicx}
\usepackage{subfigure}
\usepackage{amsmath}
\usepackage{bm}
\usepackage{dsfont}
\usepackage{mathtools}
\usepackage{nccmath}
\usepackage{amssymb}
\usepackage{multirow}
\usepackage{booktabs}
\usepackage{varwidth}
\usepackage{pifont}
\usepackage{makecell}
\usepackage{wrapfig}
\usepackage{varwidth}
\usepackage{makecell}
\usepackage{enumitem}
\usepackage{adjustbox}
\usepackage{graphicx,calc}
\usepackage{amsfonts,amsthm,bm}
\usepackage[flushleft]{threeparttable}
\usepackage{pifont}
\usepackage{ulem}
\usepackage{tabularx}
\usepackage{arydshln}
\usepackage{float}
\tcbuselibrary{breakable,theorems,skins}
\usepackage{cleveref}

\usepackage{pgfmath}
\usepackage{siunitx}

\NewEnviron{elaboration}{
\par
\begin{tikzpicture}
\node[rectangle,minimum width=0.44\textwidth] (m) {\begin{minipage}{0.44\textwidth}\BODY\end{minipage}};
\draw[thick] (m.south west) rectangle (m.north east);
\end{tikzpicture}
}

\icmltitlerunning{URAG: A Benchmark for Uncertainty Quantification in Retrieval-Augmented Large Language Models}

\begin{document}

\twocolumn[
\icmltitle{URAG: A Benchmark for Uncertainty Quantification in Retrieval-Augmented Large Language Models}



\icmlsetsymbol{equal}{*}

\begin{icmlauthorlist}
\icmlauthor{Vinh Nguyen}{equal,usl}
\icmlauthor{Cuong Dang}{equal,vt}
\icmlauthor{Jiahao Zhang}{pennst}
\icmlauthor{Hoa Tran}{fpt}
\icmlauthor{Minh Tran}{hcmus}
\icmlauthor{Trinh Chau}{uet}\\
\icmlauthor{Thai Le}{iu}
\icmlauthor{Lu Cheng}{uic}
\icmlauthor{Suhang Wang}{pennst}
\end{icmlauthorlist}

\icmlaffiliation{usl}{Uppsala University}
\icmlaffiliation{hcmus}{University of Science, VNU-HCM}
\icmlaffiliation{iu}{Indiana University}
\icmlaffiliation{fpt}{FPT Software, AI Center}
\icmlaffiliation{pennst}{The Pennsylvania State University}
\icmlaffiliation{vt}{Virginia Tech}
\icmlaffiliation{uet}{VNU University of Engineering and Technology}
\icmlaffiliation{uic}{University of Illinois at Chicago}

\icmlcorrespondingauthor{Cuong Dang}{cuongdc@vt.edu}
\icmlcorrespondingauthor{Suhang Wang}{szw494@psu.edu}

\icmlkeywords{Machine Learning, ICML}

\vskip 0.5in
]



\printAffiliationsAndNotice{\icmlEqualContribution} 

\begin{abstract}

Retrieval-Augmented Generation (RAG) has emerged as a widely adopted approach for enhancing LLMs in scenarios that demand extensive factual knowledge. However, current RAG evaluations concentrate primarily on correctness, which may not fully capture the impact of retrieval on LLM uncertainty and reliability. To bridge this gap, we introduce {\textcolor{URAGColor}{\textbf{URAG}}}, a comprehensive benchmark designed to assess the \textbf{uncertainty} of RAG systems across various fields like healthcare, programming, science, math, and general text. By reformulating open-ended generation tasks into multiple-choice question answering, URAG allows for principled uncertainty quantification via conformal prediction.  We apply the evaluation pipeline to 8 standard RAG methods,  measuring their performance through both accuracy and prediction-set sizes based on LAC and APS metrics. Our analysis shows that \textbf{\ding{182}} \textit{accuracy gains often coincide with reduced uncertainty}, but this relationship \textit{breaks under retrieval noise}; \textbf{\ding{183}} \textit{simple modular RAG methods tend to offer better accuracy–uncertainty trade-offs} than more complex reasoning pipelines; and \textbf{\ding{184}} \textit{no single RAG approach is universally reliable across domains}. We further show that \textbf{\ding{185}} \textit{retrieval depth, parametric knowledge dependence, and exposure to confidence cues can amplify confident errors and hallucinations}. Ultimately, URAG establishes a systematic benchmark for analyzing and enhancing the trustworthiness of retrieval-augmented systems. \href{https://github.com/phuvinhnguyen/URAG}{Our code is available on GitHub \faGithub}.

\end{abstract}

\section{Introduction}
\label{sec: intro}

\begin{figure}[t]
\centering\scriptsize\renewcommand\arraystretch{0.}
\setlength{\tabcolsep}{0.pt}
\begin{tabular}{cc}
\includegraphics[width=1.0\linewidth]{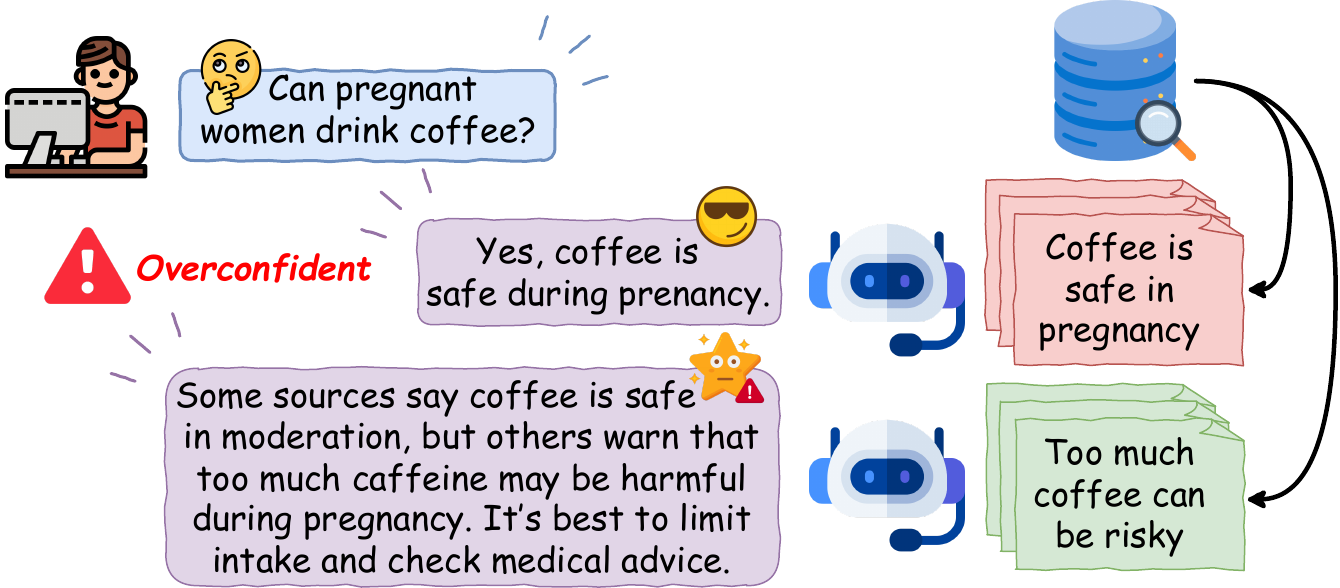}
\end{tabular}
\caption{\textbf{Illustration of uncertainty in RAG.} When asked \textit{``Can pregnant women drink coffee?''}, the retriever surfaces multiple documents, one stating that ``coffee is safe in pregnancy'' and another warning that ``too much coffee can be risky.''  The LLM, however, places disproportionate attention on the first document and produces an \textit{overconfident yet incomplete answer}, overlooking contradictory evidence. This example highlights how noisy or one-sided retrieval can amplify model overconfidence, leading to potentially harmful or misleading responses. 
}
\vskip -1em
\label{fig:rag_unc_example}
\end{figure}

Large language models (LLMs), such as GPT-5~\cite{zhang2023complete} and Claude Opus 4.1~\cite{anthropic2025claude4opus}, achieve strong performance on tasks like question answering and coding, but still struggle in knowledge-intensive domains, such as molecular prediction~\cite{xian2025molrag,sengupta2025biomol,xu2025dualequi}, autonomous driving~\cite{wei2024editable,zhang2024smartcooper,luo2025diffusion}, and personalized recommendation~\cite{zhao2024recommender,ning2024cheatagent,huang2026towards}, due to hallucinations and factual errors~\cite{ji2023towards,xu2024hallucination,wang2025knowledge}.


To address these limitations, retrieval-augmented generation (RAG)~\cite{NEURIPS2020_6b493230} grounds LLMs in external knowledge by conditioning generation on retrieved evidence, improving factual accuracy and reducing hallucinations. Recent RAG systems (e.g., REALM~\cite{pmlr-v119-guu20a}, RETRO~\cite{borgeaud2022retro}, Atlas~\cite{izacard2022atlas_arxiv}, GraphRAG~\cite{graphrag2023}) demonstrate strong performance across knowledge-intensive tasks such as open-domain QA, fact verification, and multi-hop reasoning.

However, \textit{uncertainty in RAG systems} remains a largely underexplored frontier. Although RAG can improve factual grounding, it introduces new sources of epistemic uncertainty through retrieval noise, relevance mismatch, or selective attention to partial evidence. For instance, as shown in \Figureref{fig:rag_unc_example}, when asked \textit{``Can pregnant women drink coffee?''}, the retriever may surface multiple documents, one claiming that ``coffee is safe in pregnancy'' and another warning that ``too much coffee can be risky.'' If the LLM over-relies on a single supporting document, it may generate an \textit{overconfident yet incomplete} response, potentially leading to harmful outcomes in sensitive domains. Such overconfidence in finance, law, and software engineering can result in confidently incorrect outputs and yield costly or unsafe decisions. These examples underscore the societal importance of understanding and quantifying RAG uncertainty.

Despite progress in benchmarking uncertainty for standalone LLMs~\cite{ye2024benchmarking}, the community still lacks a unified framework for assessing \textit{how retrieval reshapes model confidence and reliability}. Different RAG architectures, whether training-free, modular, or end-to-end, reasoning-oriented, may exhibit distinct uncertainty behaviors depending on the query type, retrieval relevance, and domain context. Understanding these behaviors is not only essential for developing safer and more trustworthy retrieval-augmented systems but also for improving overall accuracy by helping models better balance confidence with evidence quality. Motivated by these gaps, this work explores the following \textbf{research questions}:

\crefformat{obs}{\textcolor{mydarkgreen}{#1}}

\vspace{-0.5\baselineskip}
\keyquestion{
\vspace{-0.2\baselineskip}
\begin{itemize}[leftmargin=*, itemsep=0em]
    \item \researchquestion{\label{rq:unc_accross}How does RAG uncertainty vary across retrieval methods, domains, and prompts? (\Observations~{\hypersetup{linkcolor=mydarkgreen}\ref{obs:across_method}, \ref{obs:prompts}, \ref{obs:across_domain}}, in \Sections~\ref{subsec:ana_method}, \ref{subsec:prompt_robust}, \ref{subsec:ana_domain})}
    
    \item \researchquestion{\label{rq:acc_unc_corr}How are model accuracy and predictive uncertainty correlated across RAG systems? (\Observation~{\hypersetup{linkcolor=mydarkgreen}\ref{obs:tradeoff}} in \Sectionref{subsec:acc_unc_tradeoff})}
    \item \researchquestion{\label{rq:unc_retrieve}How does retrieval affect accuracy and predictive uncertainty under irrelevant contexts, or when the LLM already knows the answer versus when its parametric knowledge is insufficient and varying retrieval sizes? (\Observations~{\hypersetup{linkcolor=mydarkgreen}\ref{obs:irrelevant}, \ref{obs:conflict}, \ref{obs:retrieval_size}}, in \Sections~\ref{subsec:irr}, \ref{subsec:knowledge_conflict}, \ref{subsec:retrieval_size})} 
\end{itemize}
}




% ... 



Nevertheless, several \textit{technical challenges} remain in evaluating RAG uncertainty comprehensively. 
\textit{First}, there is currently \textit{no standardized pipeline} for assessing RAG uncertainty across diverse domains and architectures. 
Existing study~\cite{crag2025} focuses on accuracy or robustness but lack a unified framework for uncertainty quantification. 
To address this, we propose a \textit{domain- and RAG-agnostic pipeline} that enables fair comparison across methods without sacrificing task performance, allowing simultaneous evaluation of both accuracy and uncertainty.
\textit{Second}, \textit{measuring uncertainty in open-ended generative tasks} is inherently difficult, as free-form text outputs vary in style and reasoning chains, confounding confidence estimation. We overcome this by \textit{reformulating open-ended tasks into multiple-choice question answering (MCQA)} and leveraging \textit{conformal prediction} to obtain statistically valid uncertainty scores. However, another difficulty arises: how to generate challenging choices for MCQA. To this end, we propose a prompt and iterative generation pipeline that generates incorrect answers to be more confusing even when one knows the correct answer. \textit{Third}, \textit{isolating the contribution of retrieval to overall model uncertainty} is nontrivial because RAG systems jointly depend on the LLM’s intrinsic knowledge and the quality of retrieved evidence. 
To disentangle these factors, we \textbf{\ding{182}} evaluate RAG uncertainty with an \textit{irrelevant-retrieval database} to probe retrieval-induced epistemic noise and \textbf{\ding{183}} evaluate two complementary settings, one where the LLM already knows the answer and another where it does not, to reveal how retrieval reshapes model confidence. 
Together, these designs enable principled, cross-domain, and retrieval-aware benchmarking of RAG uncertainty.




Our findings show that retrieval does not uniformly improve reliability. While higher accuracy often matches lower uncertainty, this relationship breaks under irrelevant or noisy retrieval, especially when the LLM’s parametric knowledge is weak. We observe that simple, modular RAG methods consistently achieve better accuracy–uncertainty trade-offs than more complex aggregation or reasoning pipelines, and no single RAG method performs reliably across domains. Finally, we show that retrieval depth and confidence cues can amplify overconfident errors, revealing that retrieval can both improve performance and worsen hallucinations.

In this paper, we introduce \textbf{URAG}, the first benchmark evaluating accuracy and uncertainty across 7 RAG methods, 8 datasets over 5 domains. We further provide a pipeline to convert open-ended tasks into MCQA format.

\vspace{-0.5\baselineskip}
\contribution{
The paper's contributions can be summarized as:
\vspace{-0.2\baselineskip}
\begin{enumerate}[leftmargin=*, itemsep=0em]
    \item Introduce URAG, the first comprehensive benchmark that jointly evaluates accuracy and uncertainty in RAG methods across multiple domains.
    \item Propose a new MCQA construction pipeline that generates high-quality, challenging answer options by enforcing plausibility, semantic consistency, and confusion with the correct answer.
    \item Provide a systematic analysis across RAG methods, domains, and prompting strategies, and an in-depth study of retrieval effects, including noise, knowledge dependence, and retrieval depth, revealing that no RAG method is universally reliable and that retrieval can amplify confident errors.
\end{enumerate}
}

\vspace{-0.75\baselineskip}

\section{Related Work}
\label{sec:related_work}

\textbf{RAG Benchmarking.} 
Recent benchmarks examine complementary facets of RAG: \cite{chen2024benchmarking} probe the robustness of LLMs with RAG to noisy/irrelevant context, negative rejection, and counterfactuals \cite{chen2024benchmarking}; \emph{LaRA} \cite{lara2025} contrasts RAG with long-context routing at scale, revealing no single winning strategy across tasks; \emph{T$^{2}$\!-RAGBench} \cite{strich2025t} targets semi-structured inputs by requiring joint reasoning over text and tables; \emph{MultiHop-RAG} \cite{multihoprag2024} isolates evidence chaining for multi-step queries; \emph{RAGBench} \cite{ragbench2024} advances explainability with metrics such as TRACe to move beyond accuracy-only scoring; \emph{SafeRAG} \cite{saferag2025} stresses security under conflicting or adversarial evidence; \emph{GraphRAG-Bench} \cite{graphragbench2025} evaluates graph-structured retrieval and domain-specific reasoning; and \emph{CRAG} \cite{crag2025} offers a broad, unifying suite across tasks and retrieval scenarios. 
Beyond performance and reasoning, recent studies have also revealed the privacy vulnerabilities inherent in different retrieval architectures, covering standard RAG~\cite{zeng2024good}, Graph RAG~\cite{liu2025exposing,yang2026query,luo2026graphs}, and Multimodal RAG~\cite{al2026systemic}.
While existing works collectively map performance-accuracy, robustness, multi-hop reasoning, structure-aware retrieval, and safety, these efforts leave open how retrieval reshapes \emph{uncertainty} and calibration, and how it interacts with parametric knowledge. \textit{Our benchmark fills this gap by co-reporting task performance and predictive uncertainty, enabling principled analyses of calibration, over/under-reliance on external evidence, and the reliability of RAG-enabled systems}.

\textbf{RAG for LLMs.}
Due to space limitations, we defer the related works on RAG for LLMs to \Appendixref{appendix:related_work}.

\section{Background RAG and Uncertainty in LLMs}
\label{sec:background}

In this section, we present background about RAG and conformal prediction for RAG.

\subsection{Retrieval-Augmented Generation}
\label{sec:background_rag}
Let $\mathcal{X}$ be the user query space and $\mathcal{Y}$ be the model output space. The space of retrieval queries and the retrieval corpus are denoted by $\mathcal{Q}$ and $\mathcal{D}$, respectively. $\mathcal{A}$ is the action space governing retrieval decisions, e.g., whether to retrieve. Generally, a RAG system has four modular components.
(1) A \emph{query construction module} $G : (\mathcal{X}, \mathcal{Y}) \rightarrow \mathcal{Q}$ generates a retrieval query based on the original user query and the current model output.
(2) A \emph{retrieval policy} $\pi : (\mathcal{Q}, \mathcal{Y}) \rightarrow \mathcal{A}$ selects a retrieval action, determining whether retrieval is performed.
(3) A \emph{retriever} $R : (\mathcal{A}, \mathcal{Q}) \rightarrow \mathcal{D}$ returns a set of documents from the corpus conditioned on the selected action and query.
(4) An \emph{update function} $F : (\mathcal{X}, \mathcal{Y}, \mathcal{D}) \rightarrow \mathcal{Y}$ integrates retrieved evidence into the current output to produce an updated generation state.

\textbf{Iterative Retrieval-Generation Process.} Unlike single-shot RAG, which generates an output in a single retrieval-generation step, iterative RAG performs a retrieval-generation process over $T$ iterations, progressively refining the output through repeated interaction with the retrieval corpus.
The process is initialized with the user input query, $y_0 = x$. For $t = 1, \dots, T$, the system iteratively executes:
\begin{align}
    q_t &= G(x, y_{t-1}), \quad 
    a_t = \pi(q_t, y_{t-1}), \\
    d_t &= R(a_t, q_t), \quad
    y_t = F(x, y_{t-1}, d_t).
\end{align}

Refer \Appendixref{sec:rag_formulation_detail} for more details on how this formulation is applied in recent RAG methods.

\subsection{Conformal Prediction for Language Model}
\label{subsec:conformal}

Conformal prediction (CP) ~\cite{vovk2005algorithmic} is an emerging and theoretically-grounded topic in statistics for uncertainty quantification, especially for LLMs~\cite{ye2024benchmarking,sheng2025analyzing}. Another popular line of uncertainty quantification methods, based on entropy or perplexity, is highly sensitive to the temperature of the softmax function. Moreover, unlike conformal prediction, entropy cannot provide any guarantee on the coverage rate, because when measuring uncertainty, entropy doesn't take model accuracy into account. Entropy remains the same when predicted probabilities are permuted, even though prediction accuracy may differ. In contrast, conformal prediction (CP) provides a principled framework for quantifying predictive uncertainty with finite-sample, distribution-free guarantees. Instead of producing a single point prediction, CP outputs a prediction set that contains the true answer with high probability.

Assume that a RAG system is solving an MCQA task with option set $\mathcal{C}$. For brevity, we treat the retrieval stage implicitly. Given a query $x$, the generator, $h_\theta$ parameterized by $\theta$, induces a predictive probability vector $\text{p}=\left[p_c\right]_{c\in\mathcal{C}},\,p_c\in\left[0,1\right]$, obtained by normalizing the option-token logits, $\sum_{c\in\mathcal{C}}p_{c}=1$, as shown in \Algorithmref{alg:compute_confidence}.

CP constructs a set of candidate answers such that the true answer is included with probability at least $1-\alpha$, for a user-specified risk level $\alpha$. 
The core quantity in CP is the \emph{nonconformity score}, which measures how incompatible a candidate label is with the model’s predictive distribution for a given input $x$. Formally, given any nonconformity function $s:\mathcal{X}\times\mathcal{C}\to\mathbb{R}$, CP constructs a set predictor for $x$
\[
\mathcal{S}_\alpha(x)
=
\{\,c\in\mathcal{C}:\; s(x,c)\le \hat q_\alpha \,\},
\]
which satisfies the finite-sample coverage guarantee
\begin{equation}\label{eq:cp-coverage}
\mathbb{P}\big\{ c^\star \in \mathcal{S}_\alpha(X) \big\} \ge 1-\alpha,
\end{equation}
where $c^\star \in \mathcal{C}$ denotes the true answer and $\hat q_\alpha$ is a pre-calculated threshold according to the target risk level $\alpha$. Intuitively, small prediction sets, i.e., $\mathcal{S}_\alpha(x)$, indicate high confidence, while larger sets signal uncertainty.

In this paper, we adopt two standard methods to compute the nonconformity score: \textbf{LAC} and \textbf{APS}.

\textbf{LAC} \citep{sadinle2019least} uses
\begin{align}\label{eq:lac}
s_{\mathrm{LAC}}(x,c) &= 1-p_c(x),
\end{align}
yielding the smallest average set size among labelwise rules (but potentially under-covering hard cases).  

\textbf{APS} \citep{romano2020classification} is rank-aware, with cumulative-mass score
\begin{align}\label{eq:aps}
s_{\mathrm{APS}}(x,c) &= \sum_{c':\,p_{c'}(x)\ge p_c(x)} p_{c'}(x),
\end{align}
typically improving per-instance coverage at the cost of larger sets, especially when retrieval makes $\pi_c(x)$ diffuse (i.e., higher retrieval-induced uncertainty).

The threshold $\hat{q}_\alpha$ is the empirical $(1-\alpha)$-quantile of the calibration score $\{{s_j}\}_{j=1}^n$:
\begin{equation}\label{eq:cp-quant}
\hat{q}_\alpha
=
\mathrm{quant}\left(
\{{s_j}\}_{j=1}^n,
\frac{\lceil (n+1)(1-\alpha)\rceil}{n}
\right).
\end{equation}


\begin{figure*}[!t]
\centering\scriptsize\renewcommand\arraystretch{0.}
\setlength{\tabcolsep}{0.pt}
\begin{tabular}{cc}
\includegraphics[width=0.97\linewidth]{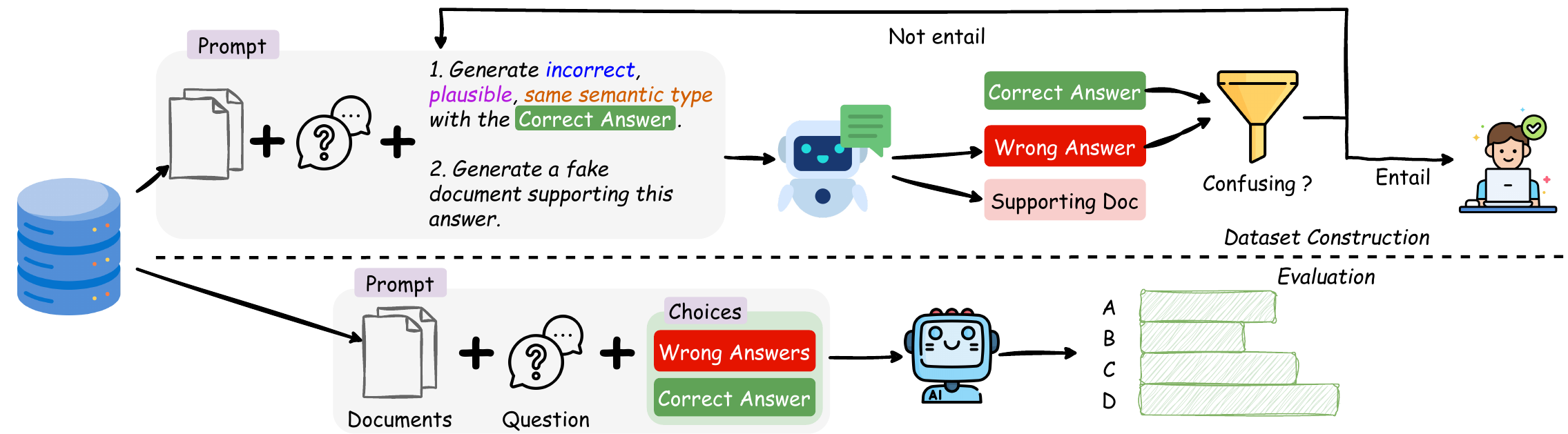}
\end{tabular}
\caption{\textbf{Illustration of dataset construction and evaluation pipeline for URAG}. The upper flow shows how we construct the benchmark, which involves retrieving documents from the database and prompting LLMs to generate a supporting document and wrong answers. After that, we use an NLI model to confirm the difficulty before having annotators check the final result. During the evaluation process, RAG systems decide between a list of wrong answers and the correct answer before measuring performance and uncertainty.}
\vskip -1em
\label{fig:benchmark}
\end{figure*}

\section{Uncertainty RAG Benchmarking}
\label{sec:benchmark}

In this section, we present the URAG Benchmark, a comprehensive framework for evaluating uncertainty in retrieval-augmented generation systems. We first outline the benchmarking pipeline, detailing how open-ended RAG datasets are transformed into multiple-choice formats to enable calibrated uncertainty measurement in \Sectionref{subsec:pipeline}. We then describe datasets in \Sectionref{subsec:datasets}, and RAG methods we use for benchmarking in \Sectionref{subsec:rag_methods}.

\subsection{Benchmarking Pipeline}
\label{subsec:pipeline}


In this section, we explain the advantages of benchmarking on the MCQA task, describe the overall benchmarking pipeline, and show the difficulty of creating a reliable wrong answer generation in the MCQA task.

\textbf{Why do we need to convert to MCQA?} Open-ended generation poses fundamental challenges for uncertainty estimation, as the output space is vast and often admits many semantically valid responses, making it difficult to define model confidence. Existing uncertainty measures, such as token likelihoods or response diversity, have been shown to correlate poorly with factual correctness in open-ended settings, especially when external context is involved. Consequently, converting the task to MCQA is a common practice for uncertainty measurement in LLMs~\cite{ye2024benchmarking, Kumar2023ConformalPW, Kapoor2024LargeLM}. This constrains the output space to a finite, well-defined set of candidates, enabling unambiguous correctness labels. It also yields more reliable uncertainty quantification through calibrated option probabilities, allowing estimates to better reflect epistemic rather than linguistic uncertainty.

\textbf{Overall Pipeline.} We first employ a retriever, i.e, \texttt{sentence-transformers/all-MiniLM-L6-v2} to gather relevant documents for each query. These are paired with prompts that guide a generator, i.e., Gemini, to produce plausible but incorrect answers, which are then combined with the correct answer to form answer choices, as illustrated in \Figureref{fig:benchmark}. These plausible but incorrect answers serve as valid distractors. 

\textbf{Valid Distractors.} 
A central challenge in constructing a reliable MCQA benchmark is preventing models from exploiting superficial cues, such as selecting options that are lexically or semantically distinct, or trivially eliminating implausible answers. To address this, we enforce explicit validity constraints on each generated distractor. In particular, a valid distractor must: \textbf{\ding{182}} be factually incorrect, \textbf{\ding{183}} remain plausible in the context of the question, and \textbf{\ding{184}} share the same semantic type as the correct answer. To further encourage plausibility, we prompt the LLM to generate a synthetic supporting document for each candidate distractor, as illustrated in \Figureref{fig:benchmark}. The full prompt is in \Promptref{prompt:fake_ans}.


\begin{table}[t]
    \centering
    \caption{
    \textbf{Progressive improvement in valid distractor generation using iterative NLI-based filtering.} Starting from a naive prompt, each iteration applies \Prompt~{\textcolor{Promptcolor}{\ref{prompt:fake_ans}}} to regenerate distractors that fail the entailment-based difficulty check.
    }
    \label{tab:difficultyprocess}
    \footnotesize
    \vskip -0.5em
    \setlength{\tabcolsep}{6pt}
    \begin{tabular}{lcccc}
        \toprule
        \textbf{Dataset} 
        & \textbf{Naive} 
        & \textbf{Iter.\,1} 
        & \textbf{Iter.\,2} 
        & \textbf{Iter.\,3} \\
        & \Promptref{prompt:fake_ans_naive}
        & \Promptref{prompt:fake_ans}
        & \Promptref{prompt:fake_ans}
        & \Promptref{prompt:fake_ans} \\
        \midrule
        LCA      & $0\%$ & $100\%$ & $100\%$ & $100\%$ \\
        CRAG     & $31\%$ & $87\%$  & $93\%$  & $96\%$ \\
        NewsSum  & $10\%$ & $98\%$  & $100\%$ & $100\%$ \\
        \bottomrule
    \end{tabular}
    \vskip -1em
\end{table}

Our design is motivated by the intuition that difficult MCQs induce residual uncertainty even after identifying the correct answer, as plausible incorrect options can be logically or semantically related to the correct answer. For example, when a human examinee solves a difficult MCQ, they may identify the correct answer but still hesitate among several plausible alternatives, resulting in reduced confidence. To operationalize this intuition, we employ a natural language inference (NLI) model to assess the difficulty and confusability of each distractor. Given the original question and an NLI model, a constructed MCQ is considered to be \textit{`challenging'} if there is \textit{at least one distractor} that is predicted to be entailed by the correct answer. In case there's no entailed incorrect answer, we regenerate many times with \Promptref{prompt:regen_fake_ans} to make it more confusing. \Table~\ref{tab:difficultyprocess} shows the percentage of `difficult' questions after each iteration using this pipeline.



\subsection{Datasets}
\label{subsec:datasets}

\Figureref{fig:taks} illustrates the tasks and datasets used to evaluate both the accuracy and uncertainty of RAG methods. This benchmark spans five diverse domains, including mathematics, general language understanding, scientific research, code generation, and healthcare.
Specifically, we evaluate on CRAG~\cite{crag2025}, NewsSum~\cite{alex2019multinews}, DialFact~\cite{gupta-etal-2022-dialfact}, SciFact~\cite{wadden-etal-2020-fact}, LCA~\cite{Bogomolov2024LongCA}, ODEX~\cite{wang2023execution}, OlympiadBench~\cite{he-etal-2024-olympiadbench}, and HealthVer~\cite{sarrouti-etal-2021-evidence-based}.
Additional dataset construction details are provided in \Appendixref{appendix:datasets} while examples and an overview of the dataset are provided in \Table~\ref{tab:datasets}, \Table~\ref{tab:full_dataset_examples}, and \Table~\ref{tab:database_statistics}.

The underlying retrieval corpora vary substantially in size and granularity across tasks.
For example, ODEX contains approximately 34K short code-related documents, while its diagnostic variant W/Odex uses a reduced corpus of 3.51K short question–answer documents.
Similarly, W/DialFact consists of 3.51K short question–answer documents derived from DialFact.
OlympiadBench includes 270K math problems with annotated solutions.
LCA is based on a single large GitHub code repository, NewsSum uses one full research paper per instance, HealthVer retrieves from an average of 3.18 medical research papers, SciFact retrieves from approximately 1.06 scientific papers, and CRAG retrieves from an average of five long documents.
Complete corpus statistics are summarized in \Tableref{tab:database_statistics}.

\subsection{RAG Methods}
\label{subsec:rag_methods}
We choose diverse RAG methods: (i) spanning from training-free, including Fusion~\cite{rackauckas2024rag}, HyDE~\cite{gao2022hyde}, RAT~\cite{wang2024rat}, Naive RAG~\cite{NEURIPS2020_6b493230}; (ii) end-to-end, including FiD~\cite{izacard2021fid} and Self-RAG~\cite{asai2024selfrag}; (iii) independently trained, including RAPTOR~\cite{sarthi2024raptor}; and sequential training, including REPLUG~\cite{shi2024replug}. Refer \Appendixref{appendix:rag_description} for more descriptive details.


\subsection{Metrics and RAG Uncertainty Quantification}
\label{subsec:metrics}
Inspired by prior work \cite{ye2024benchmarking}, we evaluate Accuracy \textbf{Acc}, Set Size \textbf{SS}, and Coverage Rate \textbf{CR}. (i) \textbf{Acc} measures the fraction of instances where the model’s top-ranked answer matches the ground-truth label. (ii) The prediction-set size \textbf{SS} reflects the width of the uncertainty set produced by conformal prediction. A larger \textbf{SS} indicates that the model assigns non-negligible probability mass to multiple answer options, revealing higher uncertainty. In this benchmark, we use the average set size based on both \textbf{LAC} and \textbf{APS} mentioned in \Sectionref{subsec:conformal}; and (iii) We also report the coverage rate to verify if the coverage guarantee requirement shown in \Equationref{eq:cp-coverage} has been satisfied. A formal definition of \textbf{Acc}, \textbf{SS}, \textbf{CR} is given in \Appendixref{appendix:metrics}.

A key challenge in evaluating RAG effectiveness arises from the intrinsic knowledge of the underlying LLM. A model equipped with stronger parametric knowledge may achieve higher accuracy regardless of the retrieval mechanism, potentially inflating performance scores and obscuring the true contribution of the RAG component. To mitigate this, we additionally report the ratio between the performance of each RAG-augmented model and its corresponding LLM-only baseline in \Tableref{tab:rag_unc_llm_8b}. This ratio reflects the degree to which retrieval enhances or hinders the model’s reasoning and uncertainty calibration.





\begin{table*}[t]
  \centering
  \footnotesize
  \setlength{\tabcolsep}{4pt}
  \caption{\textbf{Accuracy, Coverage, and Uncertainty results of different RAG methods across tasks using Llama-3.1-8B-Instruct.} \textbf{``W/o Retrieve''} denotes the baseline without retrieval. For each dataset, the top three methods in terms of highest accuracy and lowest uncertainty are highlighted in red.}
  \label{tab:rag_unc_llm_8b}
  \begin{adjustbox}{center, max width=\textwidth}
  \begin{tabularx}{.83\textwidth}{l *{8}{c}}
    \toprule
    \multicolumn{1}{l}{} & 
    \multicolumn{1}{c}{\textbf{Healthcare}} & 
    \multicolumn{2}{c}{\textbf{Code}} & 
    \multicolumn{1}{c}{\textbf{Research}} & 
    \multicolumn{1}{c}{\textbf{Math}} &
    \multicolumn{3}{c}{\textbf{General Text}} \\
    \cmidrule(lr){2-2}\cmidrule(lr){3-4}\cmidrule(lr){5-5}\cmidrule(lr){6-6}\cmidrule(lr){7-9}
    \textbf{RAG} &
    \textbf{Healthver} &
    \textbf{ODEX} &
    \textbf{LCA} &
    \textbf{SciFact} &
    \textbf{Olympiad} &
    \textbf{CRAG} &
    \textbf{NewsSum} &
    \textbf{DialFact} \\
    \midrule

    \multicolumn{9}{c}{\textit{Performance} -- \textbf{Acc (\%) $\uparrow$}} \\
    \cdashline{1-9}[2.5pt/5pt]\noalign{\vskip 0.5ex}
\textbf{W/o Retrieve} & 0.45 & 0.88 & 0.21 & 0.45 & 0.34 & 0.55 & 0.36 & 0.47 \\
\textbf{FiD} & 0.37 & 0.28 & 0.21 & 0.42 & 0.30 & 0.31 & 0.26 & 0.35 \\
\textbf{Fusion} & 0.52 & \textcolor{red}{0.86} & \textcolor{red}{0.82} & 0.69 & 0.37 & 0.66 & 0.41 & 0.71 \\
\textbf{HyDE} & 0.53 & 0.85 & 0.73 & \textcolor{red}{0.72} & 0.40 & 0.62 & \textcolor{red}{0.43} & \textcolor{red}{0.72} \\
\textbf{RAPTOR} & 0.51 & 0.85 & 0.73 & 0.70 & 0.39 & 0.67 & 0.38 & \textcolor{red}{0.72} \\
\textbf{RAT} & 0.49 & 0.84 & 0.34 & 0.65 & \textcolor{red}{0.46} & 0.67 & 0.40 & 0.64 \\
\textbf{REPLUG} & 0.51 & \textcolor{red}{0.86} & 0.73 & 0.70 & 0.36 & 0.67 & 0.38 & 0.71 \\
\textbf{Self-RAG} & 0.51 & 0.83 & 0.74 & 0.70 & 0.40 & 0.63 & 0.38 & 0.68 \\
\textbf{Naive} & \textcolor{red}{0.54} & \textcolor{red}{0.86} & 0.76 & 0.70 & 0.40 & \textcolor{red}{0.68} & 0.37 & \textcolor{red}{0.72} \\
    \midrule

    \multicolumn{9}{c}{\textit{Coverage Rate} -- \textbf{CR (\%) $\uparrow$}} \\
    \cdashline{1-9}[2.5pt/5pt]\noalign{\vskip 0.5ex}
\textbf{W/o Retrieve} & 0.90 & 0.92 & 0.91 & 0.90 & 0.89 & 0.90 & 0.87 & 0.93 \\
\textbf{FiD} & 1.00 & 0.95 & 0.94 & 1.00 & 0.95 & 0.94 & 0.99 & 0.94 \\
\textbf{Fusion} & 0.91 & 0.94 & 0.92 & 0.90 & 0.87 & 0.92 & 0.91 & 0.91 \\
\textbf{HyDE} & 0.91 & 0.93 & 0.91 & 0.89 & 0.93 & 0.91 & 0.89 & 0.90 \\
\textbf{RAPTOR} & 0.92 & 0.93 & 0.94 & 0.88 & 0.88 & 0.91 & 0.90 & 0.91 \\
\textbf{RAT} & 0.90 & 0.92 & 0.93 & 0.92 & 0.87 & 0.91 & 0.90 & 0.90 \\
\textbf{REPLUG} & 0.93 & 0.97 & 0.97 & 0.90 & 0.89 & 0.90 & 0.92 & 0.95 \\
\textbf{Self-RAG} & 0.90 & 0.92 & 0.94 & 0.93 & 0.90 & 0.92 & 0.87 & 0.91 \\
\textbf{Naive} & 0.92 & 0.92 & 0.93 & 0.86 & 0.91 & 0.92 & 0.90 & 0.90 \\
    \midrule

    \multicolumn{9}{c}{\textit{Prediction Uncertainty} -- \textbf{SS $\downarrow$}} \\
    \cdashline{1-9}[2.5pt/5pt]\noalign{\vskip 0.5ex}
\textbf{W/o Retrieve} & 2.62 & \textcolor{red}{1.66} & 4.64 & 2.59 & 3.48 & 2.61 & 2.94 & 2.55 \\
\textbf{FiD} & 3.00 & 3.63 & 4.76 & 3.00 & 3.98 & 3.69 & 3.98 & 2.84 \\
\textbf{Fusion} & 2.64 & 1.73 & \textcolor{red}{2.19} & 2.18 & 3.36 & 2.38 & 2.82 & 2.00 \\
\textbf{HyDE} & 2.49 & 1.68 & 2.38 & 2.00 & 3.69 & 2.42 & 2.74 & \textcolor{red}{1.97} \\
\textbf{RAPTOR} & 2.65 & 1.71 & 2.70 & 1.98 & 3.40 & \textcolor{red}{2.32} & 2.69 & 2.05 \\
\textbf{RAT} & 2.58 & 1.70 & 4.47 & 2.46 & \textcolor{red}{3.30} & 2.50 & 3.05 & 2.22 \\
\textbf{REPLUG} & \textcolor{red}{2.24} & 3.50 & 4.63 & 2.57 & 3.85 & 3.72 & 3.73 & 2.69 \\
\textbf{Self-RAG} & 2.69 & 1.77 & 2.59 & 2.32 & 3.51 & 2.52 & \textcolor{red}{2.66} & 2.05 \\
\textbf{Naive} & 2.65 & 1.69 & 2.46 & \textcolor{red}{1.94} & 3.54 & 2.29 & 2.77 & 2.05 \\
    \bottomrule
      \end{tabularx}
  \end{adjustbox}
\vskip -1em
\end{table*}


\section{Benchmark Results}

This section presents a comprehensive benchmark and answers \RQref{rq:unc_accross} about uncertainty insights across RAG methods in~\Sectionref{subsec:ana_method} 
and prompts in~\Sectionref{subsec:prompt_robust} and answers \RQref{rq:acc_unc_corr} about accuracy-uncertainty correlation in \Sectionref{subsec:acc_unc_tradeoff}. Due to space constraints, we defer the analysis on the impact of domains to \Appendixref{subsec:ana_domain}, and defer detailed experiment setup to \Appendixref{appendix:experiement_setup} . 

\subsection{Analysis by RAG methods} 
\label{subsec:ana_method}

From \Table~\ref{tab:rag_unc_llm_8b}, we observe distinct uncertainty patterns across RAG architectures. Modular, training-free methods, Fusion, Naive RAG, and HyDE achieve the most balanced trade-off between accuracy and reliability, maintaining high average accuracies (Acc$\approx 0.63$) with low uncertainty (SS $\approx 2.42$). RAPTOR and Self-RAG perform comparably, validating the benefits of recursive summarization and tree-structured retrieval. These methods leverage structured reasoning and controlled retrieval, improving factual grounding while preserving slightly higher uncertainty, 2.43 and 2.51, respectively. In contrast, REPLUG enhances factual accuracy in several domains but exhibits much higher uncertainty (SS $\approx 3.36$). Because REPLUG processes each passage sequentially and fuses responses through probabilistic addition, its predictions depend less on any individual passage yet amplify disagreement across them, leading to elevated epistemic variance. RAT achieves modest accuracy (Acc$\approx 0.56$) yet maintains moderate uncertainty (SS$\approx 2.78$), indicating that its multi-step retrieval pipeline 
diverts attention from the correct answers.
Finally, FiD suffers the poorest overall calibration, with the highest uncertainty (mean $SS \approx 3.61$) and lowest accuracy ($\approx 0.31$). Its reliance on a lower-capacity encoder-decoder backbone (T5) limits its ability to effectively fuse multiple passages, particularly in code and research domains, resulting in unstable uncertainty estimates.

\vspace{-0.25\baselineskip}
\begin{observation}[Answer for \textbf{\RQref{rq:unc_accross}}]\label{obs:across_method}
\textcolor{Observationcolor}{$\bullet$}~Simpler and modular pipelines (e.g., Fusion, Naive) achieve high performance with low uncertainty.
\textcolor{Observationcolor}{$\bullet$}~The augmented CoT approach, e.g., RAT, has low accuracy with high uncertainty. 
\textcolor{Observationcolor}{$\bullet$}~Sequential inference followed by probabilistic aggregation, e.g., REPLUG, has high accuracy with underconfidence.
\end{observation}
\vspace{-0.25\baselineskip}

\subsection{Brittleness to Prompts}
\label{subsec:prompt_robust}

\begin{figure}[!ht]
\centering\scriptsize\renewcommand\arraystretch{0.}
\setlength{\tabcolsep}{0.pt}
\begin{tabular}{cc}
\includegraphics[width=0.97\linewidth]{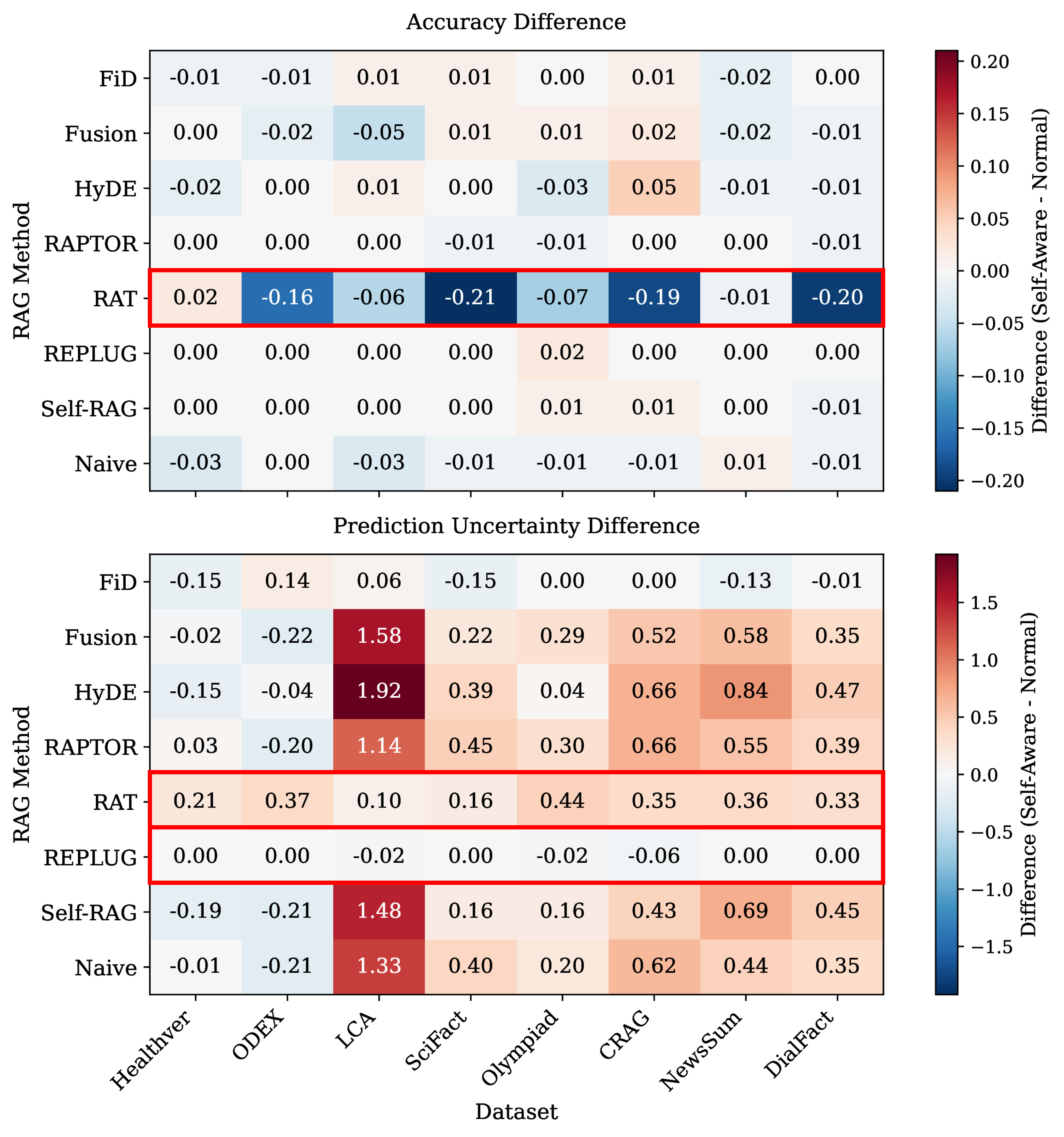}
\end{tabular}
\vskip -1em
\caption{\textbf{Accuracy and uncertainty difference under self-aware prompting.} When the LLM is exposed to its confidence scores, RAT exhibits the strongest degradation in accuracy and stability, while most other RAG methods show reduced uncertainty, particularly on math (Olympiad) and general-domain tasks (CRAG).}
\label{fig:self-aware}
\vskip -1.2em
\end{figure}

\textbf{Self-aware Evaluation.}
\cite{boldt2019confidence} shows that human confidence can be influenced by prior self-assessments when solving similar tasks. Following it, we introduce an experiment, the \textit{Self-Aware Evaluation}, by using \Promptref{prompt:mcqa-wrong-aware}. Specifically, the model is provided with its own confidence scores obtained from an initial forward pass. This enables us to examine whether explicitly exposing the model to its internal belief state influences its decision-making process and to what extent retrieval affects this behavior. \Figureref{fig:self-aware}, which compares results of \Tableref{tab:rag_unc_llm_8b} and \Tableref{tab:rag_unc_llm_8b_self-aware}, shows that when being exposed to its confidence, while RAT is affected with a decrease in accuracy and an increase in uncertainty across almost all tasks, other methods stay unaffected in terms of performance. However, uncertainty rises on most benchmarks, decreasing only on HealthVer and ODEX. Lastly, we observe that the performance and uncertainty of REPLUG are almost immutable across all benchmarks.

\begin{figure}
    \centering
    \includegraphics[width=0.97\linewidth]{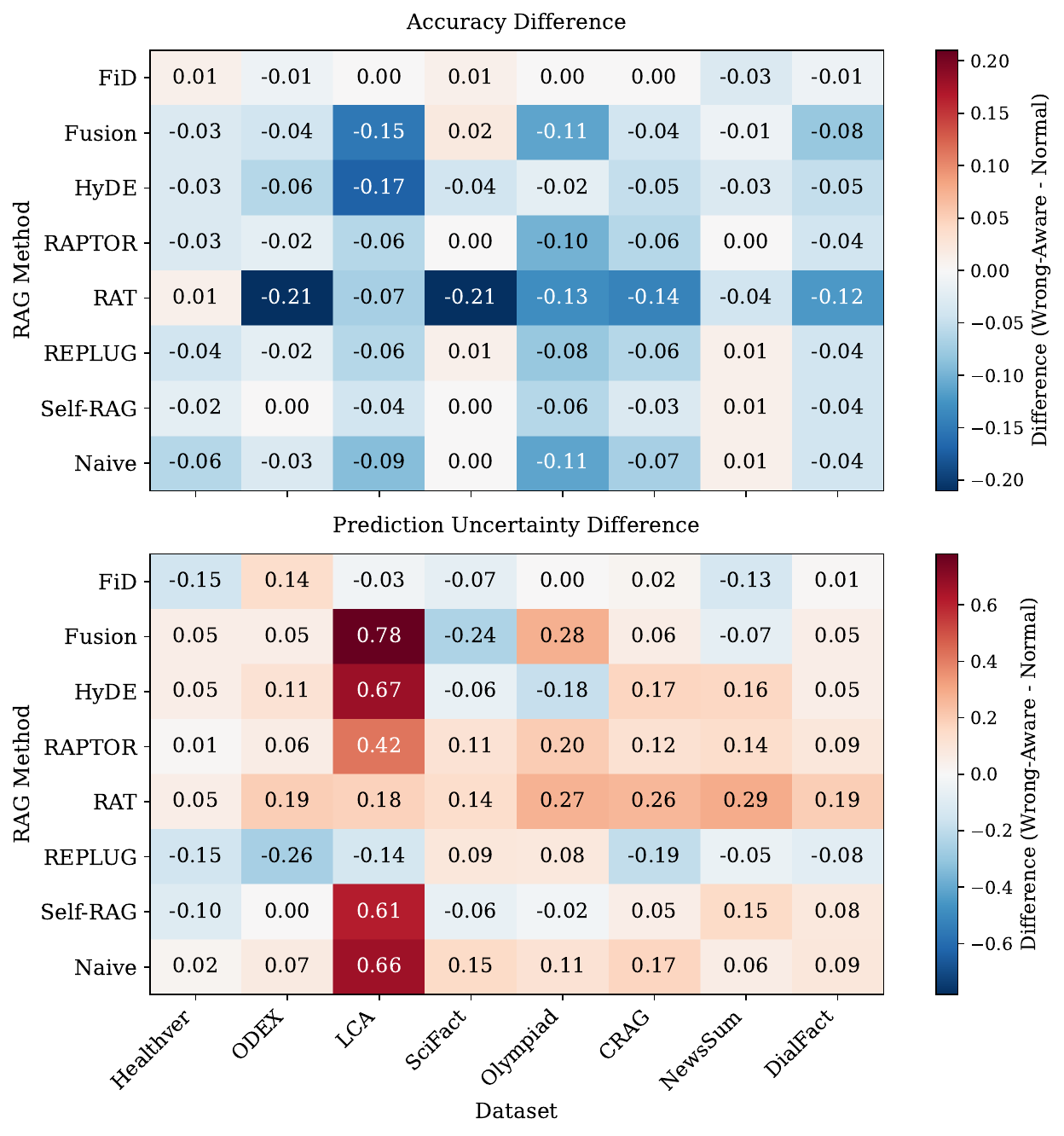}
    \vskip -1em
    \caption{\textbf{Accuracy and uncertainty difference under wrong-aware prompting.} Misleading confidence cues cause a consistent drop in accuracy and an increase in uncertainty, indicating a higher rate of hallucinated decisions.}

\label{fig:wrongaware_heatmap_highlight}
\end{figure}

\textbf{Wrong-aware Evaluation.}
In the self-aware setting, the model’s predicted answer and the revealed confidence distribution were aligned, which resulted in only marginal changes across most RAG systems. To further probe the model’s reliance on externally presented confidence signals, we introduce a wrong-aware evaluation. This experiment preserves the same prompt structure as the self-aware setup but deliberately perturbs the confidence distribution by swapping the highest confidence option with the lower one. This manipulation examines whether the system prioritizes retrieved evidence over misleading belief cues. Notably, we do not modify factual content, retrieved documents, or answer options; only the reported confidence levels are altered. These results are reported in \Figureref{fig:wrongaware_heatmap_highlight} and \Tableref{tab:rag_unc_llm_8b_wrong_aware}. We observe consistent degradation in both accuracy and certainty across most RAG methods, indicating that misleading confidence cues substantially increase hallucinated behavior. From a security perspective, the wrong-aware prompt can be interpreted as a form of prompt injection that induces hallucinations by manipulating the model’s perceived belief state rather than its retrieved evidence.

\vspace{-0.25\baselineskip}
\begin{observation}[Answer for \textbf{\RQref{rq:unc_accross}}]\label{obs:prompts}\textcolor{Observationcolor}{$\bullet$}Exposing a model to its own confidence scores systematically alters its uncertainty behavior without affecting performance, except for RAG.
\textcolor{Observationcolor}{$\bullet$}~REPLUG is largely invariant to confidence-based prompt perturbations.
\textcolor{Observationcolor}{$\bullet$}~RAG systems struggle to prioritize retrieved evidence over deceptive belief cues.
\end{observation}
\vspace{-0.25\baselineskip}


\begin{figure}[!ht]
\centering\scriptsize\renewcommand\arraystretch{0.}
\setlength{\tabcolsep}{0.pt}
\vskip -1em
\begin{tabular}{cc}
\includegraphics[width=0.87\linewidth]{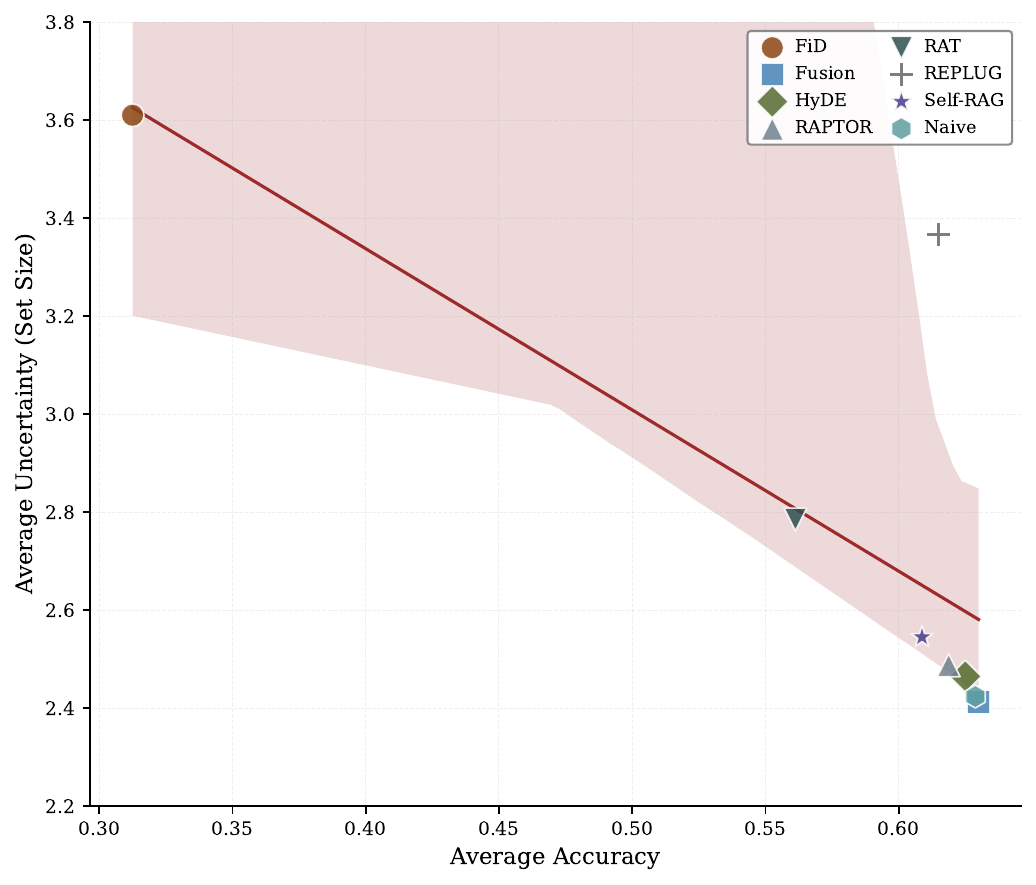}
\end{tabular}
\caption{\textbf{Reverse Correlation between Accuracy and Uncertainty across RAG methods.} 
Each point represents the average accuracy and predictive uncertainty (set size) of one RAG method evaluated on ten datasets. 
A strong negative correlation ($r=-0.76$) shows that higher accuracy aligns with lower predictive uncertainty across RAG methods.}
\vskip -1em
\label{fig:acc_unc}
\end{figure}

\subsection{Accuracy-Uncertainty Correlation}
\label{subsec:acc_unc_tradeoff}

\Figure~\ref{fig:acc_unc} shows an inverse relationship between accuracy and uncertainty across eight tasks, with REPLUG as an exception. This deviation arises because REPLUG processes each retrieved document independently and aggregates their predicted probabilities, which reduces reliance on any single passage and mitigates overconfident predictions.

\vspace{-0.25\baselineskip}
\begin{observation}[Answer for \textbf{\RQref{rq:acc_unc_corr}}]\label{obs:tradeoff}
Almost all RAG variants, higher accuracy coincides with lower uncertainty.
\end{observation}
\vspace{-0.25\baselineskip}

\begin{figure}[!ht]
\centering\scriptsize\renewcommand\arraystretch{0.}
\setlength{\tabcolsep}{0.pt}
\vskip -1em
\begin{tabular}{cc}
\includegraphics[width=0.97\linewidth]{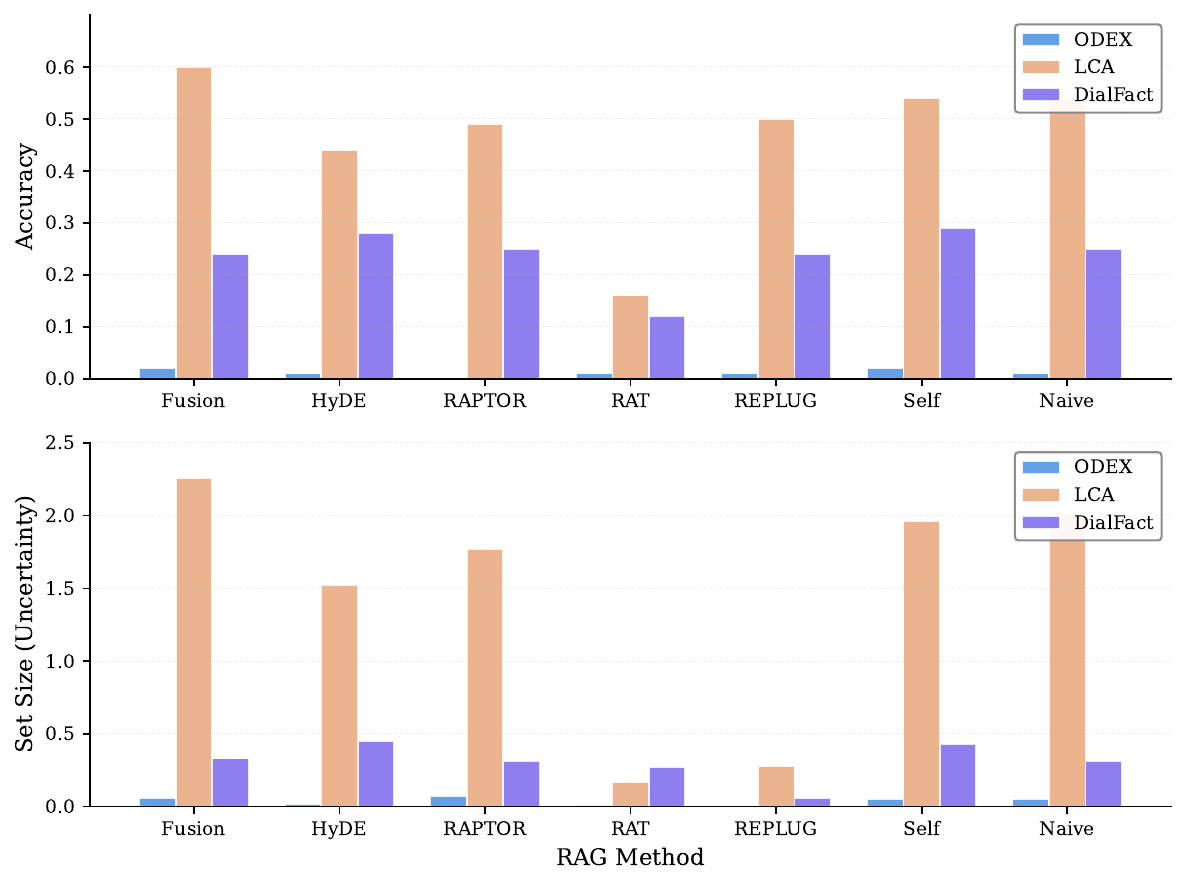}
\end{tabular}
\vskip -1em
\caption{\textbf{Impact of Irrelevant Contexts on Accuracy and Uncertainty.} We add irrelevant information along with retrieved information on DialFact, Odex, and LCA. Bars show absolute differences between with/without noisy information.}
\vskip -1em
\label{fig:irr_context}
\end{figure}

\section{Dissecting Retrieval Effects: Irrelevance, Knowledge Isolation, and Retrieval Size}

In this section, we answer \RQref{rq:unc_retrieve} by further analyzing the performance and uncertainty of LLMs when dealing with irrelevant contexts in \Section~\ref{subsec:irr} and benchmarking the RAGs on questions that LLM answer correctly/incorrectly in \Section~\ref{subsec:knowledge_conflict}. 
Due to space limitations, we defer the observation on the effect of the retrieval size to \Appendixref{subsec:retrieval_size}. 



\subsection{Uncertainty Irrelevant Contexts}
\label{subsec:irr}

We simulate the scenario when retrieved information is noisy by randomly collecting 10 irrelevant documents from the database and adding them to the original 10 retrieved documents every time the RAG method retrieves information from the database. The evaluation is conducted on three subsets: ODEX, LCA, and DialFact. The experimental results are reported in \Figureref{fig:irr_context} and \Tableref{tab:rag_results_reorganized}. The results show that task that base LLM already performs well, i.e. ODEX, will not be affected by noisy information. However, LCA and DialFact observed a strong degradation in the accuracy and certainty of most RAG methods, highlighting potential LLM ability degradation in handling irrelevant information. 

\vspace{-0.25\baselineskip}
\begin{observation}[Answer for \textcolor{RQcolor}{\textbf{\RQref{rq:unc_retrieve}}}]\label{obs:irrelevant}
\textcolor{Observationcolor}{$\bullet$}~RAG becomes less sensitive to irrelevant context when the underlying LLM already possesses strong parametric knowledge.
\end{observation}
\vspace{-0.25\baselineskip}


\subsection{Knowledge Isolation for RAG Benchmarking}
\label{subsec:knowledge_conflict}

\begin{figure}[!ht]
\centering\scriptsize\renewcommand\arraystretch{0.}
\setlength{\tabcolsep}{0.pt}
\vskip -1em
\begin{tabular}{cc}
\includegraphics[width=0.97\linewidth]{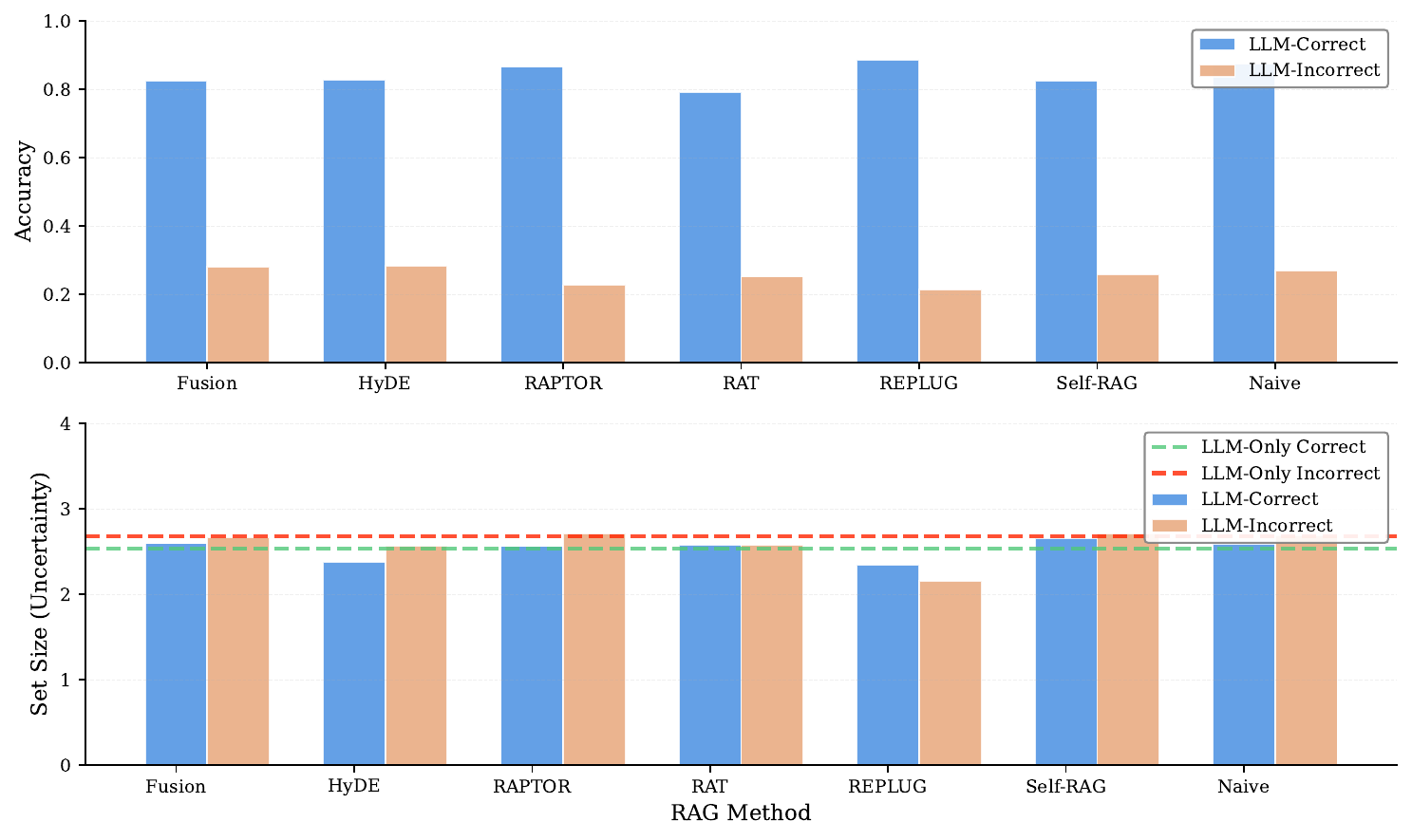}
\end{tabular}
\vskip -0.5em
\caption{\textbf{Comparison of RAG accuracy and uncertainty on LLM-Correct and LLM-Incorrect cases.} ``LLM-Only Correct'' and ``LLM-Only Incorrect'' in the below chart show the Set Size of LLM (w/o retrieval) in two sets. The charts report mean values on the HealthVer benchmark. The extended results are in \Table~\ref{tab:rag_correct_incorrect_8b}.}
\vskip -1.5em
\label{fig:conflict}
\end{figure}


In this section, we analyze retrieval effects under two complementary conditions: questions that the LLM can already answer correctly and questions that it cannot.

Specifically, we first evaluate the LLM-only baseline and divide the test set into two disjoint groups: \textbf{LLM-Correct}, consisting of questions that the LLM answers correctly using only its parametric knowledge, and \textbf{LLM-Incorrect}, consisting of questions that the LLM answers incorrectly. We then apply RAG methods to each subset separately.

The \textbf{LLM-Correct} set enables us to examine whether introducing external evidence alters the model’s original, already-correct decisions and how retrieval affects confidence when the answer is known. In contrast, the \textbf{LLM-Incorrect} set captures cases where the LLM’s parametric knowledge is insufficient, allowing us to study how retrieval contributes to accuracy improvements and reshapes predictive uncertainty.

\Figureref{fig:conflict} show that RAPTOR and REPLUG achieve higher accuracy on the LLM-Correct set but lower accuracy on the LLM-Incorrect set, indicating that these methods place relatively less reliance on retrieved evidence and instead defer more strongly to the LLM’s parametric knowledge.

HyDE and REPLUG also increase model confidence on the LLM-Correct set. While applying RAG improves accuracy on the LLM-Incorrect set, the associated predictive uncertainty remains unchanged across different RAG methods.

\vspace{-0.25\baselineskip}
\begin{observation}[Answer for \textbf{\RQref{rq:unc_retrieve}}]\label{obs:conflict}
\textcolor{Observationcolor}{$\bullet$}~RAPTOR and REPLUG rely less on retrieval.\textcolor{Observationcolor}{$\bullet$}~HyDE and REPLUG increase certainty when the LLM already knows the answer.
\end{observation}
\vspace{-0.25\baselineskip}

\section{Conclusion}

This work introduced URAG, a unified benchmark for evaluating RAG accuracy and uncertainty via conformal prediction. 
By reformulating open-ended RAG tasks into MCQA and leveraging conformal prediction, URAG enables principled, statistically grounded uncertainty quantification across diverse domains and RAG architectures. URAG establishes a strong foundation for future research into trustworthy retrieval-augmented LLM systems.


\section*{Impact Statement}

This work presents a benchmark for evaluating the uncertainty of Retrieval-Augmented Generation (RAG) methods using conformal prediction. By aiming to enhance the reliability and trustworthiness of generative systems, this research contributes to safer AI deployment. We do not foresee any negative societal impacts from this work.

\bibliographystyle{icml2026}
\bibliography{ref}

\onecolumn
\clearpage
\section*{\Large{Appendix}}
\setcounter{section}{0}
\setcounter{figure}{0}
\setcounter{table}{0}
\makeatletter 
\renewcommand{\thesection}{\Alph{section}}
\renewcommand{\theHsection}{\Alph{section}}
\renewcommand{\thefigure}{A\arabic{figure}}
\renewcommand{\theHfigure}{A\arabic{figure}}
\renewcommand{\thetable}{A\arabic{table}}
\renewcommand{\theHtable}{A\arabic{table}}
\makeatother

\renewcommand{\thetable}{A\arabic{table}}
\setcounter{mylemma}{0}
\renewcommand{\themylemma}{A\arabic{mylemma}}
\setcounter{algorithm}{0}
\renewcommand{\thealgorithm}{A\arabic{algorithm}}
\setcounter{equation}{0}
\renewcommand{\theequation}{A\arabic{equation}}

\section{Extended Related Works}\label{appendix:related_work}

\textbf{RAG for LLMs.} 
We view retrieval-augmented generation (RAG) for LLMs through five complementary design patterns that differ in how retrieval is incorporated and trained, and, crucially for our goals, in the kinds of uncertainty signals they expose. In \emph{training-free} systems, retrieval is introduced at inference without updating model parameters, typically by augmenting prompts with retrieved passages or by conditioning token generation on nearest neighbors. Prompt-level augmentation exemplified by HyDE synthesizes a hypothetical passage from the query to guide retrieval and then conditions the LLM on the top evidence \cite{gao2022hyde,cuong-etal-2024-curious}, whereas token-level conditioning, as in kNN-LM \cite{khandelwal2020knnlm} and RETRO \cite{borgeaud2022retro}, injects neighbors or chunked memories directly into the generative process. These approaches are attractive for their simplicity and model-agnostic deployment, yet their predictive confidence often hinges on prompt budget, retrieval noise, and format sensitivity, making them fertile ground for calibration analyses. 

Recent extensions pursue adaptive retrieval strategies that expose richer uncertainty signals while maintaining compatibility with frozen language models. REPLUG \cite{shi2024replug} prepends retrieved documents directly to black-box LM inputs without specialized cross-attention, enabling application to any existing model including proprietary systems; critically, the LM supervises the retriever through its own prediction signals, creating a bidirectional feedback loop where retrieval adapts to generation needs while the generator itself remains frozen, offering uncertainty estimates through retriever confidence and document selection patterns. RAPTOR \cite{sarthi2024raptor} constructs multi-level document trees by recursively applying UMAP dimensionality reduction and Gaussian Mixture Model clustering to text chunks, then summarizing each cluster via LLM to form parent nodes in a bottom-up hierarchy; at inference, the collapsed tree is queried via cosine similarity across all abstraction levels simultaneously, exposing uncertainty through the distribution of retrieved nodes across hierarchical depths and the semantic coherence of multi-granularity evidence. Chain-of-Retrieval Augmented Generation (CoRAG) \cite{corag2024} enables iterative multi-step reasoning by \emph{training} language models to decompose questions into sub-query sequences, retrieve evidence for each sub-query, and synthesize intermediate sub-answers before final generation; the training procedure employs rejection sampling over candidate retrieval chains to augment datasets without manual annotation, and at test time, strategies such as best-of-N sampling with penalty scoring or breadth-first tree search allow the model to explore multiple reasoning paths and surface uncertainty through chain diversity, conditional likelihoods of retrieval failure, and learned stopping mechanisms that predict when sufficient information has been gathered.

Moving beyond purely inference-time integration, \emph{independently trained} pipelines learn the retriever and the generator separately and combine them only at inference, supporting modular upgrades and clearer attribution of error. Dense Passage Retrieval (DPR) \cite{karpukhin2020dpr} paired with a frozen reader and Fusion-in-Decoder (\textsc{FiD}) \cite{izacard2021fid} as a strong multi-passage reader are canonical examples. The decoupling makes retriever similarity scores and reader fusion behavior natural uncertainty hooks, allowing us to probe how confidence tracks with passage relevance and diversity. A related but more interdependent pattern is \emph{sequential training}, where one component is trained first and the other adapted while the first is fixed. REALM pretrains a retriever before training the generator \cite{pmlr-v119-guu20a}; conversely, “LLM-first” variants tune the generator on clean data and then train a retriever to surface compatible contexts. The stagewise nature of these methods introduces a potential mismatch between components, offering a controlled axis to study uncertainty arising from distributional shifts at the interface.

At the other end of the spectrum, \emph{joint} or end-to-end training optimizes the retriever and generator together—either by marginalizing over retrieved evidence or by backpropagating through retrieval decisions. RAG-Sequence and RAG-Token \cite{NEURIPS2020_6b493230} instantiate this paradigm, and Atlas \cite{izacard2022atlas_arxiv} scales it to large-corpus pretraining with retrieval in the loop. Because these models induce a posterior over documents and tokens, they naturally support uncertainty estimates via evidence marginalization and sampling-based disagreement, enabling principled calibration analyses tied to the retrieval distribution itself. Finally, an increasingly important thread targets \emph{structured retrieval for reasoning}, extending beyond unstructured text to graphs, tables, tools, or explicit retrieval policies. GraphRAG integrates graph-structured context into generation, leveraging paths and neighborhoods as evidence \cite{graphrag2023}, while Self-RAG equips models with a mechanism to decide \emph{when} and \emph{what} to retrieve during multi-step reasoning \cite{asai2024selfrag}. The explicit structure and policies in these systems surface additional, fine-grained sources of uncertainty, from path reliability and cell provenance to policy confidence about retrieval actions.

Across these families, the community has made rapid progress on accuracy and efficiency, yet the pathways by which retrieval reshapes LLM uncertainty, both epistemic and aleatoric, and the calibration of downstream predictions remain underexplored. Our benchmark is designed to make these pathways observable by aligning evaluation protocols with method-specific “uncertainty hooks”: retriever scores and top-$k$ diversity for training-free and independent pipelines, interface mismatch for sequential designs, evidence posteriors for jointly trained models, and structure- or policy-aware signals for reasoning-oriented retrieval. By reporting uncertainty quality alongside task performance within a unified framework, we aim to clarify not only \emph{whether} RAG helps, but \emph{how} different integration choices modulate confidence, reliability, and robustness.

\section{Experiment Setup}
\label{appendix:experiement_setup}
We implement a range of RAG methods, including Fusion, HyDE, RAPTOR, RAT, REPLUG, Self-RAG, and Naive RAG, using LLaMA-3.1-8B\footnote{{\href{https://huggingface.co/meta-llama/Llama-3.1-8B}{meta-llama/Llama-3.1-8B}}} as the backbone generator. The only exception is FiD, which employs the T5\footnote{\href{https://huggingface.co/Intel/fid_flan_t5_base_nq}{Intel/fid\_flan\_t5\_base\_nq}} model due to its encoder–decoder architecture. For document embeddings, we use the Sentence Transformer\footnote{\href{https://huggingface.co/sentence-transformers/all-MiniLM-L6-v2}{sentence-transformers/all-MiniLM-L6-v2}}.

As the generator, we use LLaMA-3.1-8B or LLaMA-3.2-3B-Instruct, with a temperature of 0.1 to reduce stochasticity, improve reproducibility, and facilitate fair comparisons across methods. Unless otherwise specified, we report results using LLaMA-3.1-8B. For all tasks, the retrieval depth is fixed at 10 documents. The benchmarking results for LLaMA-3.1-8B are shown in \Tableref{tab:rag_unc_llm_8b}, and for LLaMA-3.2-3B-Instruct are shown in \Tableref{tab:rag_unc_llm_3b}.

We adopt Naive RAG and set the conformal calibration risk level to $\alpha = 0.1$ when computing both APS and LAC thresholds across tasks. All MCQA evaluations are conducted using \Promptref{prompt:mcqa}.

To evaluate and analyze RAG methods, we use three metrics, including accuracy, coverage rate, and set size. While the main metrics, accuracy and set size, are used for performance and uncertainty evaluation, coverage rate is used to verify the core guarantee of conformal prediction. The coverage rate should be roughly $0.9$, matching our chosen risk level $\alpha=0.1$.

\section{Datasets}
\label{appendix:datasets}

\begin{figure}[!t]
\centering\scriptsize\renewcommand\arraystretch{0.}
\setlength{\tabcolsep}{0.pt}
\begin{tabular}{cc}
\includegraphics[width=.7\linewidth]{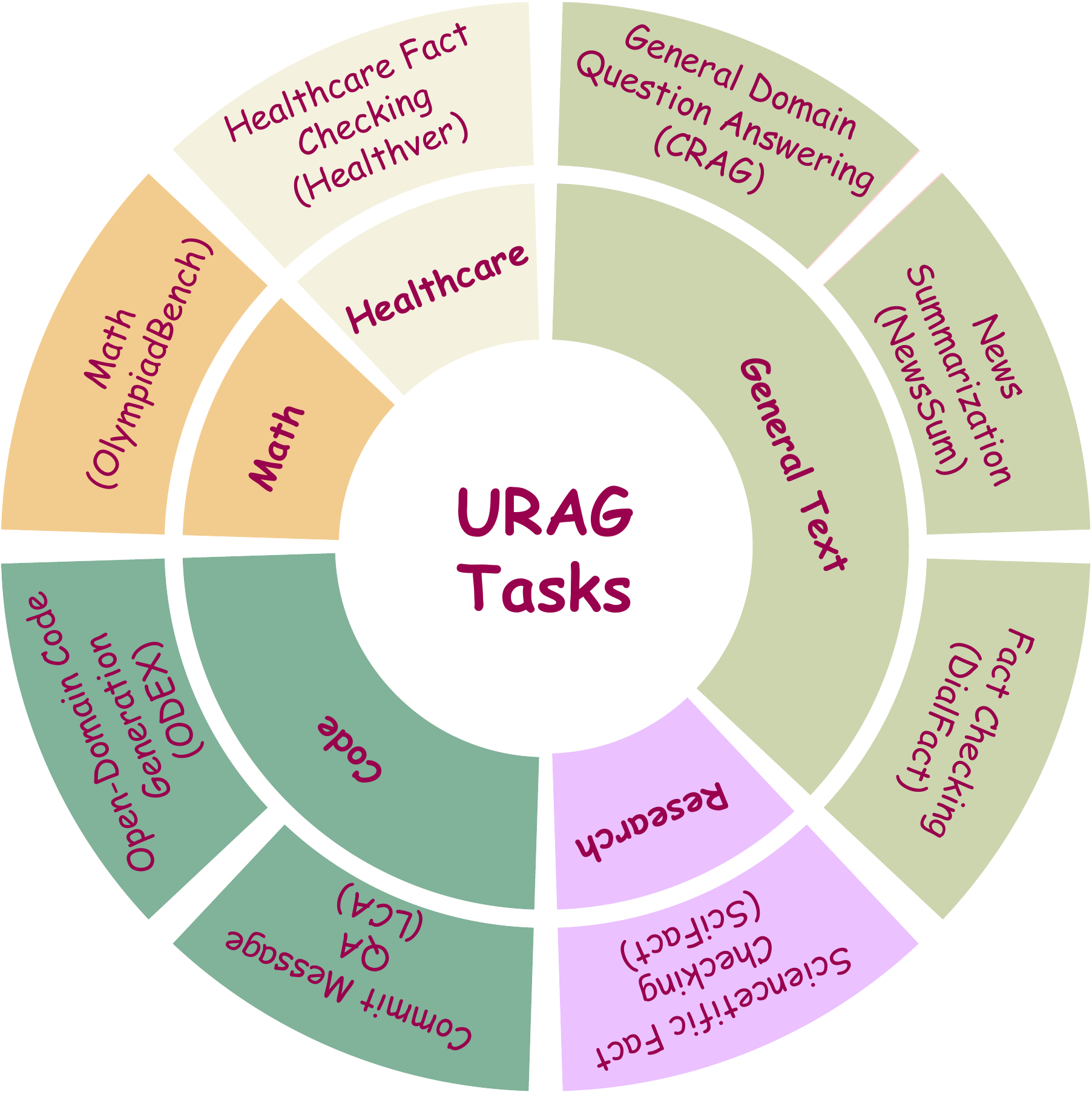}
\end{tabular}
\caption{\textbf{Tasks covered by URAG.} URAG includes tasks spanning 5 domains, general text, research, code, math, and healthcare, for examining the accuracy and uncertainty in daily-use/abstractive-reasoning/high-stakes tasks.}
\label{fig:taks}
\end{figure}

\begin{table*}[htpb]
\centering
\caption{\textbf{Overview of RAG Benchmark Datasets.}}
\label{tab:datasets}
\resizebox{\textwidth}{!}{
\begin{tabular}{lp{6cm}ccc}
\toprule
\textbf{Dataset} & \textbf{Description} & \textbf{Calibration} & \textbf{Test} & \textbf{Total} \\
\midrule
Commit Message QA & Evaluating RAG system performance in project understanding & 81 & 82 & 163 \\
\midrule
CRAG\textsuperscript{\textdagger} & Evaluating RAG system performance across multiple domains & 1,181 & 1,149 & 2,330 \\
\midrule
DialFact & Multiple choice QA subset for dialogue fact verification & 1,000 & 1,000 & 2,000 \\
\midrule
HealthVer & Health claim verification in MCQA format & 665 & 667 & 1,332 \\
\midrule
Multi-NewsSum & MCQA generated from Multi-News summarization dataset & 475 & 475 & 950 \\
\midrule
ODEX & Evaluating RAG systems in simple coding tasks & 220 & 219 & 439 \\
\midrule
OlympiadBench & Evaluating RAG systems in solving mathematics problems & 332 & 329 & 661 \\
\midrule
SciFact & Scientific claim verification in MCQA format & 187 & 187 & 374 \\
\midrule
ODEX (Wrong Context) & ODEX variant with incorrect context for robustness testing & 220 & 219 & 439 \\
\midrule
DialFact (Wrong Context) & DialFact variant with incorrect context for robustness testing & 200 & 200 & 400 \\
\bottomrule
\multicolumn{5}{l}{\small \textsuperscript{\textdagger}CRAG domains: Sports (439), Finance (552), Open (474), Music (332), Movie (533)} \\
\multicolumn{5}{l}{\small \textsuperscript{\textdagger}CRAG question types: Simple (637), Multi-hop (203), Comparison (293), Aggregation (280),} \\
\multicolumn{5}{l}{\small \quad Set (220), Simple w/ Condition (343), False Premise (265), Post-processing (89)} \\
\end{tabular}
}
\end{table*}

To comprehensively evaluate RAG uncertainty across diverse knowledge settings, we curate eight datasets spanning major domains such as code, mathematics, scientific research, news summarization, fact-checking, and healthcare. Each dataset is transformed into MCQA format. This unified design enables consistent measurement of predictive uncertainty from model outputs, independent of textual variation in open-ended responses.

Beyond domain diversity, we further introduce two diagnostic variants, Wrong Odex and Wrong DialFact, that deliberately pair questions with irrelevant or mismatched retrieval contexts. These datasets are specifically constructed to probe how RAG systems behave under retrieval noise, examining whether they can recognize misleading evidence or instead exhibit overconfidence when the retrieved information is incorrect. The summarization and example of each dataset are presented in \Table~\ref{tab:datasets} and \Table~\ref{tab:full_dataset_examples}. The followings describe how we constructed each dataset for our evaluation.
 
\textbf{CRAG.} In this benchmark, we used the existing CRAG benchmark (which was used to evaluate RAG methods) to construct correct question-answer pairs. However, to evaluate the uncertainty, we must convert the original dataset to an MCQA dataset. The conversion method is illustrated in Figure~\ref{fig:benchmark}, where we first retrieve some documents from the corpus and ask language models (Gemini) to generate wrong answers based on the question. Wrong answers are then checked before merging with the golden answer to form a multiple-choice answer set.

\textbf{Olympiad bench.}
Following a similar conversion methodology as used for CRAG, we generate plausible but incorrect answers to form multiple-choice questions. The generated distractors are carefully designed to maintain consistent formatting with the correct solution while incorporating subtle logical or computational errors. As a result, distinguishing the correct answer requires multi-step reasoning and precise mathematical computation, making this dataset particularly effective for probing how retrieval influences model confidence and uncertainty in high-difficulty reasoning tasks.

\textbf{SciFact.} The SciFact dataset consists of scientific claims extracted from peer-reviewed research papers, intending to determine whether each claim is supported, refuted, or not sufficiently supported by the accompanying evidence documents. In our benchmark, each question is reformulated as a three-choice multiple-choice task reflecting these possible verdicts. The retrieval corpus comprises a curated collection of research papers and abstracts containing semantically related claims, ensuring that the RAG model must identify and reason over precise textual evidence to reach a correct conclusion. This setup enables systematic evaluation of how retrieval quality and evidence selection influence model confidence and uncertainty in scientific fact-verification scenarios.

\textbf{Healthver.} The HealthVer dataset focuses on fact-checking within the medical and healthcare domain, ensuring that each claim can be validated against reliable evidence drawn from verified scientific and clinical sources. Similar to SciFact, each instance involves determining whether a medical claim is supported, refuted, or not sufficiently supported by the retrieved evidence. Owing to this structural similarity, we follow the same MCQA conversion process used for SciFact, reformulating each claim–evidence pair into a three-choice question. This dataset provides a rigorous setting for assessing how RAG systems handle domain-specific uncertainty in high-stakes, evidence-sensitive contexts such as medical reasoning and health information verification.

\textbf{LCA Commit Message.} This dataset is designed to evaluate how RAG systems assist LLMs in understanding and summarizing code changes. Each instance consists of a modification made to a GitHub repository, represented by the diff text alongside the corresponding source code before and after the change. The task is to find the correct commit message that accurately summarizes the modification. A gold-standard commit message is provided for each example to serve as the correct answer, where wrong answers are collected from other samples in the dataset that share high similarity with the BM25 score.

\textbf{ODEX.} The ODEX dataset evaluates RAG systems on code generation and functional reasoning tasks. Each instance provides a partially written Python function, and the objective is to find the missing code segment that correctly solves the problem. The retrieval database includes documentation and code examples from various Python libraries commonly used by LLMs, ensuring that each question reflects realistic programming contexts. While the golden answer is provided in the original dataset, false answers are retrieved with a similar method to the LCA Commit Message dataset.

\textbf{Multinew summary.} The Multi-News dataset consists of multi-document news articles paired with human-written summaries. These distortions encompass various types of factual and logical errors, including hallucination (introducing unverified information), contradiction (asserting claims that conflict with evidence), partial truth (omitting key details to yield incomplete conclusions), wrong conclusion (drawing incorrect inferences from correct facts), exaggeration (overstating or downplaying evidence), conflation (mixing up entities or events), temporal error (misrepresenting the timeline of events), and scope error (such as shifting from local to global claims). Each generated summary is carefully verified before inclusion in the benchmark to ensure reliability. This formulation enables systematic evaluation of how retrieval-augmented models handle factual grounding, reasoning accuracy, and confidence calibration in complex summarization tasks.

\textbf{Dialfact.} The DialFact dataset focuses on fact-checking within conversational contexts, containing crowd-annotated claims drawn from dialogue exchanges and paired with supporting or refuting evidence from Wikipedia. The task is to determine whether a conversational claim is supported, refuted, or lacks sufficient information based on the retrieved context. To preserve the realism of dialogue-based reasoning, both the conversational history and relevant evidence passages are provided to the model. Due to its similarity to Scifact and Healthver, we use the same method to construct this dataset.

\textbf{Wrong Odex.}
In this variant, we deliberately modify the original Odex MCQA dataset by substituting its programming-oriented retrieval database with an unrelated corpus drawn from the healthcare domain. This design serves to evaluate how RAG systems respond when retrieval introduces misleading or semantically irrelevant information. The objective is to assess the model’s ability to detect and disregard inappropriate context, as well as to examine whether it exhibits overconfidence when relying on incorrect evidence.

\textbf{Wrong DialFact.}
Similarly, in this variant, we modify the original DialFact MCQA dataset by replacing its dialogue-based evidence repository with documents sourced from the healthcare domain. The resulting mismatch between question and context enables an analysis of how RAG systems manage irrelevant retrieval in conversational fact-checking scenarios. Specifically, this variant evaluates whether the system can recognize inconsistent evidence or whether it maintains high confidence despite grounding its responses in unrelated information.

\begin{table*}[htpb]
\centering
\caption{\textbf{Dataset Examples with Full Questions and Options}. Wrong DialFact and Wrong Odex share similar question-answer, but we use a different database.}
\label{tab:full_dataset_examples}
\begin{tabular}{p{2cm}p{13cm}p{1.2cm}}
\toprule
\textbf{Category} & \textbf{Full Question with Options} & \textbf{Answer} \\
\midrule
\textbf{Commit Message QA} & 
Given the context, which commit message best describes the following code changes? 
A. Improve speed of rebalance script - This removes the call to \texttt{nodetool ring}, which can get unreasonably slow as the amount of data in a BOP cluster increases. It also adds a couple flags that allow the \texttt{nodetool status} call to be skipped if the user is already sure the sanity checks will pass. 
B. Reduce pressure on memory in stream tests - This change runs the python garbage collector before and after each stream test. The garbage collector is disabled in the CI since it has a significant impact on the duration of the jobs. 
C. Refactor FFmpegSource - Using 2 queues for video packets and audio packets. Whenever the queues have space, more packets are read from the stream. This work will allow to remove more easily the audio thread in favor of a scheduled call to refill the audio player. 
D. Support CUDA stream on memory pool - Now, memory pool will have an arena (bins) for each stream to avoid concurrent streams touch the same memory block &
D \\
\hline

\textbf{Multinewsum} &
Which of the following best summarizes the given document?
A. Today's 11 gubernatorial races are primarily focused on local issues, with Democrats poised for significant gains, possibly taking control of a majority of state offices...
B. Republicans already control the North Carolina governorship, having wrested it from Democratic control. They also solidified their hold on Utah, North Dakota, and Indiana...
C. Today's gubernatorial elections are proving to be a difficult day for Republicans, who are struggling to hold onto their seats in Utah, North Dakota, and Indiana...
D. It's a race for the governor's mansion in 11 states today, and the GOP could end the night at the helm of more than two-thirds of the 50 states. The GOP currently controls 29 of the country's top state offices... &
D \\
\hline


\textbf{OlympiadBench} &
Let $n$ be a positive integer and fix $2 n$ distinct points on a circumference. Split these points into $n$ pairs and join the points in each pair by an arrow (i.e., an oriented line segment). The resulting configuration is good if no two arrows cross, and there are no arrows $\overrightarrow{A B}$ and $\overrightarrow{C D}$ such that $A B C D$ is a convex quadrangle oriented clockwise. Determine the number of good configurations.
Options:
A. $C_n = \frac{1}{n+1}\binom{2n}{n}$
B. $\frac{(2n)!}{2^n n!}$
C. $\binom{2n}{2} \binom{2n-2}{2}...\binom{2}{2}$
D. $\binom{2n}{n}$ &
D \\
\hline

\textbf{DialFact} &
Given the conversation context, evaluate this claim: "Well, it's a little different. but if you're looking for something different, you can try gluten-free pasta. it's wheat-flavored and comes in a variety of textures and shapes. i really like the rice flour pasta"
What is the verification status?
A. Supports
B. Refutes
C. Not Enough Information &
B \\
\hline

\textbf{Healthver} &
Can taking medication to lower fever, such as paracetamol (tylenol) and ibuprofen (advil) worsen COVID-19?
A. Supported
B. Refuted
C. Not Enough Information &
A \\
\hline

\textbf{Scifact} &
Is the following scientific claim supported by evidence?
Claim: The risk of female prisoners harming themselves is ten times that of male prisoners.
A. Supported
B. Refuted
C. Not Enough Information &
A \\
\hline

\textbf{CRAG} &
What are the names of all the movies in the Chronicles of Narnia franchise?
A. The Lion, the Witch \& the Wardrobe, The Horse and His Boy, and The Magician’s Nephew
B. The Lion, the Witch \& the Wardrobe, The Voyage of the Dawn Treader, and The Last Battle
C. The names of the movies in the Chronicles of Narnia franchise are "The Lion, the Witch, and the Wardrobe", "Prince Caspian", and "The Voyage of the Dawn Treader".
D. The Lion, the Witch \& the Wardrobe, Prince Caspian, and The Silver Chair &
C \\
\hline

\textbf{ODEX} &
check if all elements in list `myList` are identical def f\_3844801(myList): return
A. [mydict[x] for x in mykeys]
B. all(x == myList[0] for x in myList)
C. list2 = [x for x in list1 if x]
D. all(predicate(x) for x in string) &
B \\

\bottomrule
\end{tabular}
\end{table*}

\section{Technical Details}

\subsection{Details on RAG Formulation}
\label{sec:rag_formulation_detail}

This unified formulation in \Sectionref{sec:background_rag} subsumes several representative RAG variants. In \textsc{HyDE}~\cite{gao2022hyde}, the query construction module $G$ generates hypothetical documents whose embeddings are used for retrieval. In \textsc{RAT}~\cite{wang2024rat} and \textsc{Self-RAG}~\cite{asai2024selfrag}, $G$ reformulates the query using intermediate generation traces or partial outputs. In standard single-shot RAG~\cite{NEURIPS2020_6b493230}, $G$ reduces to the identity mapping and the loop executes for a single step. In \textsc{RAPTOR}~\cite{sarthi2024raptor}, the retriever $R$ performs tree-traversal over hierarchically clustered, LLM-summarized indices, enabling the policy $\pi$ to retrieve multi-level, coarse-to-fine evidence that supports the update function $F$.

The policy $\pi$ is instantiated explicitly in \textsc{Self-RAG}, where a dedicated model predicts whether retrieval is necessary. The update function $F$ also varies across methods: traditional RAG conditions generation on concatenated retrieved documents, \textsc{REPLUG}~\cite{shi2024replug} and \textsc{Self-RAG} processes retrieved passages independently, and \textsc{RAT} explicitly revises prior generation traces using newly retrieved evidence.

\subsection{Extended Algorithms}

Here, we present the algorithm to compute the LLM answer probabilities for MCQAs. 

\begin{algorithm}[!ht]
\caption{Compute LM Answer Probability for MCQAs}
\label{alg:compute_confidence}
\begin{algorithmic}[1]
\STATE \textbf{Input:} User query $x$, answer options $\mathcal{C}$, generator $h_\theta$, position of answer token $i$
\STATE $\mathbf{z} \gets \left[h_\theta(x)\right]_{i}$ \hfill // logits over vocabulary at answer token
\STATE $\mathbf{z} \gets \left[\, \mathbf{z}\left[c\right] : c \in \mathcal{C} \,\right]$ \hfill // logits over options
\STATE $\mathbf{p} \gets \mathrm{Softmax}\left(\mathbf{z}\right)$
\STATE \textbf{Output:} answer probability vector $\mathbf{p}$
\end{algorithmic}
\end{algorithm}

\subsection{Metrics}
\label{appendix:metrics}
Let the test set be $\mathcal{B}_{\text{test}}={(x_i, c_i^\star)}_{i=1}^{n}$, where $x_i$ is the input query, $c_i^\star \in \mathcal{C} = {1,\dots,K}$ is the ground-truth class,  $\hat{c}(x_i)$ is the predicted class, $\mathcal{S}(x)\subseteq \mathcal{C}$ is the conformal prediction set of classes produced for input $x$.

\paragraph{Accuracy (Acc).}
Accuracy is the fraction of instances for which the RAG model’s top-ranked class matches the ground-truth class 
\[
\mathbf{Acc}
=
\frac{1}{|\mathcal{B}_\text{test}|}
\sum_{(x_i,c_i^\star)\in\mathcal{B}_\text{test}}
\mathbf{1}\!\left(\hat{c}(x_i)=c_i^\star\right).
\]

\paragraph{Set Size (SS).}
Set Size measures the average size of the conformal prediction class set
\[
\mathbf{SS}
=
\frac{1}{|\mathcal{B}_\text{test}|}
\sum_{(x_i,c_i^\star)\in\mathcal{B}_\text{test}}
|\mathcal{C}(x_i)|.
\]

\paragraph{Coverage Rate (CR).}
Coverage evaluates whether conformal prediction satisfies the coverage guarantee
\[
\mathbf{CR}
=
\frac{1}{|\mathcal{B}_\text{test}|}
\sum_{(x_i,c_i^\star)\in\mathcal{B}_\text{test}}
\mathbf{1}\!\left(c_i^\star\in\mathcal{C}(x_i)\right).
\]
Accuracy and uncertainty can trade off in non-intuitive ways: models with higher accuracy may exhibit higher uncertainty. By reporting CR, the benchmark ensures that all models are compared under the same reliability constraint. Otherwise, a model could artificially reduce SS by violating coverage, leading to misleading rankings.

\subsection{RAG Setup}
\label{appendix:rag_description}

\paragraph{FiD (Fusion-in-Decoder).}
Fusion-in-Decoder or \textsc{FiD} instantiates retrieval-augmented generation with a decoder-centric evidence fusion mechanism. Given a user query $x$, a dense retriever identifies a set of relevant documents
\(
d = R(a, x) \subset \mathcal{D},
\)
which serve as external evidence for answer generation. At the reading stage, the model takes as input the question together with the retrieved support passages and generates the answer. Retrieved documents are segmented into passages $\{p_1,\ldots,p_k\}$. Each passage, together with its associated title, is concatenated with the original query. Special tokens (\texttt{question:}, and \texttt{context:}) are prepended to the corresponding fields to distinguish their roles within the input sequence. The encoder produces passage-level representations.
\[
h_i = E(x, p_i), \quad i = 1,\ldots,k,
\]
where $E$ denotes the encoder applied to a question-passage pair. Evidence integration is deferred to the decoding stage. The final output is generated by conditioning on the collection of passage representations,
\[
y = F\!\big(x, h_1 \oplus \cdots \oplus h_k\big),
\]
where $\oplus$ denotes concatenation of passage-level representations, forming a unified encoder memory over which the decoder performs joint cross-attention during autoregressive generation. In this work, we use the pretrained \texttt{Intel/fid\_flan\_t5\_base\_nq} model. By processing passages independently in the encoder while performing joint attention in the decoder, \textsc{FiD} scales to a large number of retrieved contexts, as self-attention is applied to one passage at a time. Consequently, the computational complexity grows linearly with the number of passages rather than quadratically. At the same time, decoder-side fusion enables effective aggregation of complementary evidence from multiple passages.

\paragraph{Fusion RAG.}
\textsc{Fusion RAG} generates multiple diverse retrieval queries from a single user query, enabling retrieval to be conditioned on complementary query formulations rather than a single query. Given a user query $x$, the model constructs a set of diverse queries $\mathcal{Q} = \{q_1, q_2, \ldots, q_n\}$, where $q_1 = x$ is the original query and $\{q_2, \ldots, q_n\} = G(x)$ are alternative formulations intended to capture different phrasings or semantic aspects of the question. Diverse queries are generated using randomly selected system prompts and user instructions that encourage rephrasing, simplification, specificity, or semantic expansion, as illustrated in Figure~\ref{prompt:fusion-rag-query}. To maximize diversity, high-temperature decoding ($\tau = 0.9$) with sampling is employed during query generation.

Each query $q_i \in \mathcal{Q}$ is issued to a dense retriever, which returns an ordered list of candidate documents.
\[
\mathcal{D}_i = R(a, q_i),
\]
where documents in $\mathcal{D}_i$ are ranked by their relevance to $q_i$. The retrieval results from all queries are then fused using Reciprocal Rank Fusion (RRF), which aggregates evidence across query-specific rankings. For a document $d$ that appears at rank $r_i(d)$ in the ranked list $\mathcal{D}_i$ for query $q_i$, its fused score is defined as
\[
\mathrm{RRF}(d) = \sum_{i : d \in \mathcal{D}_i} \frac{1}{k + r_i(d) + 1},
\]
where $k$ is a smoothing constant that controls the contribution of lower-ranked documents. Documents are then sorted by their RRF scores to form a fused retrieval set $d$, which is provided to the update function $F$. To ensure a consistent input length for generation, the fused retrieved documents are concatenated and truncated to a maximum of 4{,}000 tokens before being passed to the generator. The final output is generated by conditioning on the original query and the fused retrieved context,
\[
y = F(x, y_0, d).
\]

\paragraph{HyDE (Hypothetical Document Embeddings)} 
\textsc{HyDE} generates a hypothetical document from the user query and uses it as the retrieval query, rather than embedding the original query directly. Given a user query $x$, the model generates
\[
q = G(x),
\]
where $q$ is a short, self-contained passage that hypothetically addresses the query using plausible, domain-consistent language. Hypothetical documents are generated using a fixed prompt (Figure~\ref{prompt:hyde}) that instructs the model to produce comprehensive and informative passages answering the given question. The model generates these passages solely from its parametric knowledge, without access to external documents. The hypothetical document $q$ is then issued to a dense retriever, which returns a set of candidate documents from the corpus $\mathcal{D}$. The retrieved documents $d = R(a, q)$ are combined with the original query to generate the output
\(
y = F(x, y_0, d).
\)

\paragraph{Self-RAG.}
\textsc{Self-RAG} employs a pretrained language model that produces retrieval control tokens, answer content, and reflection signals within an iterative RAG process. In our implementation, we decouple this unified model into two pretrained components, one responsible for generating retrieval control tokens ($\pi_{\text{ret}}$) and another for answer generation and reflection ($\pi_{\text{gen}}$) - to facilitate modular control and analysis. Given a user query $x$ and previous output $y_{t - 1}$, the \emph{retrieval policy} $\pi_{\text{ret}}$ decides whether to retrieve at iteration $t$:
\[
a_t = \pi_{\text{ret}}(x, y_{t - 1}) \in \{\langle|\text{retrieve}|\rangle,\langle|\text{no\_retrieve}|\rangle\}.
\]
If $a_t = \langle|\text{no\_retrieve}|\rangle$, the system proceeds without retrieval, relying solely on the LLM's parametric knowledge to generate the answer. If $a_t = \langle|\text{retrieve}|\rangle$, the retriever obtains up to five passages
$d_t = R(x) \subset \mathcal{D}$. Conditioned on $d_t$, the generator $\pi_{\text{gen}}$ produces a pool of candidate continuations $\{y_t^{(1)}, \ldots, y_t^{(M)}\}$ under different context configurations (no context, top-3 passages, or all passages). In this paper, we use a pretrained model $\pi_{\text{gen}}$ to generate answers with reflection tokens (i.e., \texttt{selfrag/selfrag\_llama2\_7b}). The generation process is formalized as: 
\[
y_t^{(m)} = \pi_{\text{gen}}(x, y_{t-1}, d_t^{(m)}),
\]
where $d_t^{(m)}$ denotes the context configuration for candidate $m$. Each candidate $y_t^{(m)}$ carries reflection tokens for relevance, support, and utility. We interpret these as scores
\[
s_{\text{rel}}^{(m)},\quad s_{\text{sup}}^{(m)},\quad s_{\text{use}}^{(m)} \in [0,1],
\]
obtained by mapping tokens such as $\langle|\text{relevant}|\rangle$ (or $\langle|\text{irrelevant}|\rangle$) to $1.0$ (or $0.0$), $\langle|\text{fully\_supported}|\rangle$ (or $\langle|\text{partially\_supported}|\rangle$, $\langle|\text{no\_support}|\rangle$) to $1.0$ (or $0.7$, $0.0$), and $\langle|\text{utility:1--5}|\rangle$ to values in $[0.2, 1.0]$ scaled by $i/5$ for utility level $i$. When reflection tokens are absent, default scores of $0.5$ are used for relevance and support. These signals are aggregated into a composite score
\[
S^{(m)} = w_{\text{rel}}\, s_{\text{rel}}^{(m)} + w_{\text{sup}}\, s_{\text{sup}}^{(m)} + w_{\text{use}}\, s_{\text{use}}^{(m)},
\]
and then select the best candidate:
\[
y_t = \arg\max_{m} S^{(m)}.
\]

In this work, we set $t = 1$ (single iteration), where $\pi_{\text{ret}}$ makes one retrieval decision and $\pi_{\text{gen}}$ selects the best candidate based on reflection token scores to produce the final output $y_t$.

\paragraph{REPLUG} augments a \emph{black-box} LM with a plug-and-play retriever, without modifying LM parameters. Given an input context $x$, a dual-encoder retriever scores each document $d$ by cosine similarity
\[
s(d,x)=\cos(E(d),E(x)),
\]
and retrieves the top-$k$ set $D'$. Instead of concatenating all retrieved texts into one prompt, \textsc{REPLUG} runs the LM separately on each concatenation $d\circ x$ and ensembles the next-token distribution:
\[
p(y \mid x, D')=\sum_{d\in D'} p(y \mid d\circ x)\,\lambda(d,x),\qquad
\lambda(d,x)=\frac{\exp(s(d,x))}{\sum_{d'\in D'}\exp(s(d',x))}.
\]
Optionally, \textsc{REPLUG-LSR} tunes the retriever using LM supervision by matching the retrieval likelihood
\[
P_R(d\mid x)=\frac{\exp(s(d,x)/\gamma)}{\sum_{d'\in D'}\exp(s(d',x)/\gamma)}
\]
to an LM-induced target distribution
\[
Q(d\mid x,y)=\frac{\exp(P_{LM}(y\mid d,x)/\beta)}{\sum_{d'\in D'}\exp(P_{LM}(y\mid d',x)/\beta)},
\]
minimizing $\frac{1}{|B|}\sum_{x\in B}\mathrm{KL}\!\left(P_R(\cdot\mid x)\,\|\,Q(\cdot\mid x,y)\right)$ while keeping the LM frozen.

\paragraph{RAPTOR (Recursive Abstractive Processing for Tree-Organized Retrieval)} indexes a corpus with a hierarchical tree whose nodes capture text at multiple levels of abstraction. It first chunks each document into short segments (100 tokens) and embeds them with SBERT, forming leaf nodes. It then iteratively applies (i) soft clustering over embeddings using Gaussian Mixture Models (after UMAP reduction) and (ii) LLM summarization within each cluster to create parent nodes; summaries are re-embedded and the process repeats until further clustering is infeasible.

Formally, for an embedding $x$, a GMM defines
\[
p(x\mid k)=\mathcal{N}(x;\mu_k,\Sigma_k),\qquad
p(x)=\sum_{k=1}^{K}\pi_k\,\mathcal{N}(x;\mu_k,\Sigma_k),
\]
and selects $K$ via Bayesian Information Criterion
\[
\mathrm{BIC}=\ln(N)\,p-2\ln(\hat{L}),
\]
where $N$ is the number of segments, $p$ is the number of model parameters, and $\hat{L}$ is the maximized likelihood.

At inference time, \textsc{RAPTOR} retrieves nodes by cosine similarity between the query embedding and node embeddings, using either (i) \emph{tree traversal} (top-$k$ selection layer-by-layer) or (ii) \emph{collapsed tree} (flatten all nodes and select the highest-scoring nodes up to a token budget; reported to perform better). The retrieved node texts are provided to the generator as context, i.e., $y=F(x,y_0,d)$.


\section{Additional Results}

 \subsection{Analysis by Domain}
 \label{subsec:ana_domain}

Because healthcare is a high-stakes domain where reliability is paramount, systems must prioritize caution, i.e., high uncertainty, especially when accuracy is low. Hence, Naive RAG is the most suitable RAG method because, as shown in \Tableref{tab:rag_unc_llm_8b}, Naive RAG not only has the highest performance but also increased caution. For \textit{Code} datasets such as ODEX, retrieval minimally affects model performance, which is also observed in previous work \citet{wang-etal-2025-coderag}. Furthermore, all RAG methods slightly raise the uncertainty of LLMs. In the \textit{Research} domain, HyDE achieves the highest accuracy (0.72) with a low uncertainty (2.00), demonstrating that hypothesis-driven retrieval effectively supports abductive reasoning. By generating intermediate hypothetical statements before retrieval, the model retrieves more causally relevant evidence, thereby improving both factual grounding and confidence. For the \textit{Math} domain, RAT attains the highest performance with the greatest certainty, confirming that iterative chain-of-thought (CoT) refinement with dynamically updated retrieval enhances precision in tasks requiring explicit reasoning steps. This design allows the system to progressively refine intermediate solutions and maintain consistent confidence.

\vspace{-0.25\baselineskip}
\begin{observation}[Answer for \textbf{\RQref{rq:unc_accross}}]\label{obs:across_domain}
\textcolor{Observationcolor}{$\bullet$}~Naive RAG is the most suitable method for the healthcare domain.
\textcolor{Observationcolor}{$\bullet$}~In domain code reasoning, retrieval has minimal impact on accuracy or uncertainty.
\textcolor{Observationcolor}{$\bullet$}~Hypothesis-based retrieval (HyDE) helps research task.
\textcolor{Observationcolor}{$\bullet$}~Iterative CoT refinement with continuous retrieval (RAT) strengthens performance with reduced uncertainty on mathematical reasoning tasks.
\end{observation}
\vspace{-0.25\baselineskip}


\subsection{Effect of the Retrieval Size}
\label{subsec:retrieval_size}

\Tableref{tab:k_sweep_prominent} shows the performance of FiD, Fusion RAG, and HyDE, on OlympiadBench, Commit Message, CRAG, and ODEX of our benchmark to illustrate the effect of the number of retrieved documents $k$. From the results, while the performance and uncertainty of HyDE and Fusion have small variances in ODEX, indicating that the extra retrieved context does not affect tasks that are well-performed by LLMs, FiD is less confident in its answers but maintains its low performance. On OlympiadBench, HyDE’s performance degrades markedly as the retrieval size $k$ increases, accompanied by a substantial rise in uncertainty. This occurs because HyDE performs retrieval based on generated hypothetical documents, which makes additional retrieved evidence increasingly noisy as $k$ grows. Lastly, the results of the LCA show three distinct patterns for the three systems as $k$ increases. While FiD remains similar, HyDE is more confident in its answer despite the insignificant change in performance, indicating this method retrieves documents supporting its answer. The performance and certainty of Fusion reduce, suggesting that irrelevant documents are retrieved as $k$ increases, causing confusion or conflicting information.

\begin{observation}[Answer for \textbf{\RQref{rq:unc_retrieve}}]\label{obs:retrieval_size}
    \textcolor{Observationcolor}{$\bullet$}~Additional retrieval does not affect both performance and uncertainty when the LLM already performs well.
    \textcolor{Observationcolor}{$\bullet$}~Larger retrieval size does not benefit the RAG system in either performance or certainty.
    \textcolor{Observationcolor}{$\bullet$}~Retrieval based on regenerated query, i.e., HyDE, is harmful in the math domain when the retrieval size increases.
\end{observation}

\begin{table}[t]
\centering
\footnotesize
\setlength{\tabcolsep}{4pt}
\caption{\textbf{Effect of retrieval depth $k$ on accuracy and uncertainty (SS) across
representative datasets from different domains.} Bold value indicates the best result on the benchmark for each RAG method.}
\label{tab:k_sweep_prominent}
\begin{tabular}{l c cc cc cc cc}
\toprule
\multirow{2}{*}{} & \multirow{2}{*}{$k$} 
& \multicolumn{2}{c}{\textbf{Olympiad}} 
& \multicolumn{2}{c}{\textbf{LCA}} 
& \multicolumn{2}{c}{\textbf{CRAG}}
& \multicolumn{2}{c}{\textbf{ODEX}} \\
\cmidrule(lr){3-4}\cmidrule(lr){5-6}\cmidrule(lr){7-8}\cmidrule(lr){9-10}
 &  & Acc & SS & Acc & SS & Acc & SS & Acc & SS \\
\midrule
\multirow{4}{*}{\rotatebox{90}{\textbf{FiD}}} & 10 & \textbf{0.30} & \textbf{3.98} & 0.21 & \textbf{4.76} & \textbf{0.31} & \textbf{3.69} & \textbf{0.28} & \textbf{3.63} \\
& 50 & 0.30 & 3.98 & 0.22 & 4.82 & 0.31 & 3.70 & 0.26 & 3.78 \\
& 100 & 0.30 & 4.00 & 0.22 & 4.79 & 0.31 & 3.69 & 0.26 & 3.76 \\
& 500 & 0.28 & 3.99 & \textbf{0.23} & 4.79 & 0.31 & 3.70 & 0.26 & 3.75 \\
\midrule
\multirow{4}{*}{\rotatebox{90}{\textbf{Fusion}}} & 10 & \textbf{0.37} & \textbf{3.36} & \textbf{0.82} & 2.19 & 0.66 & 2.38 & \textbf{0.86} & 1.73 \\
& 50 & 0.36 & 3.57 & 0.74 & \textbf{2.16} & \textbf{0.68} & \textbf{2.29} & 0.85 & \textbf{1.68} \\
& 100 & 0.36 & 3.59 & 0.71 & 2.24 & 0.68 & 2.29 & 0.85 & 1.70 \\
& 500 & 0.35 & 3.53 & 0.68 & 2.48 & 0.68 & 2.29 & 0.84 & 1.71 \\
\midrule
\multirow{4}{*}{\rotatebox{90}{\textbf{HyDE}}}& 10 & \textbf{0.40} & 3.69 & 0.73 & 2.38 & 0.62 & 2.42 & \textbf{0.85} & \textbf{1.68} \\
& 50 & 0.36 & \textbf{3.57} & \textbf{0.82} & 2.24 & \textbf{0.65} & \textbf{2.36} & 0.82 & 1.70 \\
& 100 & 0.31 & 3.71 & 0.73 & 2.18 & 0.64 & 2.44 & 0.84 & 1.73 \\
& 500 & 0.24 & 3.83 & 0.73 & \textbf{2.16} & 0.65 & 2.40 & 0.84 & 1.71 \\
\bottomrule
\end{tabular}
\end{table}

\section{Full Version of Main Empirical Results}

\Table~\ref{tab:rag_unc_llm_8b_self-aware} presents accuracy, coverage, and uncertainty of RAG methods in the Self-Aware Evaluation setting, where models receive their own predicted confidence distribution from a prior forward pass. This setup measures how RAG systems alter behavior when explicitly exposed to their internal belief states. Results cover all 8 datasets plus the two Irrelevant-Context variants. Subscripts denote LLM-calibrated uncertainty. This table highlights which RAG systems become more cautious, stable, or brittle when self-monitoring is introduced.

\begin{table*}[t]
  \centering
  \footnotesize
  \setlength{\tabcolsep}{4pt}
  \caption{\textbf{Accuracy, Coverage, and uncertainty results of different RAG methods across tasks using 8B LLM in Self-Aware Evaluation Setting (models are provided with their own confidence scores)}. The number indicates the uncertainty of the whole RAG, while the subscripts indicate the uncertainty calibrated by LLM uncertainty. This is for showing the effect of contexts on LLM.}
  \label{tab:rag_unc_llm_8b_self-aware}
  \begin{adjustbox}{center, max width=\textwidth}
  \begin{tabularx}{\textwidth}{l *{10}{c}}
    \toprule
    \multicolumn{1}{l}{} & 
    \multicolumn{1}{c}{\textbf{Healthcare}} & 
    \multicolumn{2}{c}{\textbf{Code}} & 
    \multicolumn{1}{c}{\textbf{Research}} & 
    \multicolumn{1}{c}{\textbf{Math}} &
    \multicolumn{3}{c}{\textbf{General Text}} &
    \multicolumn{2}{c}{\textbf{Irrelevant Contexts}} \\
    \cmidrule(lr){2-2}\cmidrule(lr){3-4}\cmidrule(lr){5-5}\cmidrule(lr){6-6}\cmidrule(lr){7-9}\cmidrule(lr){10-11}
    \textbf{RAG} &
    \textbf{Healthver} &
    \textbf{Odex} &
    \textbf{LCA} &
    \textbf{SciFact} &
    \textbf{Olympiad} &
    \textbf{CRAG} &
    \textbf{NewsSum} &
    \textbf{DialFact} &
    \textbf{W/DialFact} &
    \textbf{W/Odex} \\
    \midrule

    \multicolumn{11}{c}{\textit{Performance} -- \textbf{Acc (\%) $\uparrow$}} \\
\cdashline{1-11}[2.5pt/5pt]\noalign{\vskip 0.5ex}
\textbf{W/o Retrieve} & 0.45 & 0.88 & 0.21 & 0.45 & 0.37 & 0.60 & 0.37 & 0.46 & 0.43 & 0.87 \\
\textbf{FiD} & 0.36 & 0.27 & 0.22 & 0.43 & 0.30 & 0.32 & 0.24 & 0.35 & 0.33 & 0.26 \\
\textbf{Fusion} & 0.52 & 0.84 & 0.77 & 0.70 & 0.38 & 0.68 & 0.39 & 0.70 & 0.34 & 0.87 \\
\textbf{HyDE} & 0.51 & 0.85 & 0.74 & 0.72 & 0.37 & 0.67 & 0.42 & 0.71 & 0.31 & 0.88 \\
\textbf{RAPTOR} & 0.51 & 0.85 & 0.73 & 0.69 & 0.38 & 0.67 & 0.38 & 0.71 & 0.34 & 0.87 \\
\textbf{RAT} & 0.51 & 0.68 & 0.28 & 0.44 & 0.39 & 0.48 & 0.39 & 0.44 & 0.37 & 0.77 \\
\textbf{REPLUG} & 0.51 & 0.86 & 0.73 & 0.70 & 0.38 & 0.67 & 0.38 & 0.71 & 0.34 & 0.87 \\
\textbf{Self-RAG} & 0.51 & 0.83 & 0.74 & 0.70 & 0.41 & 0.64 & 0.38 & 0.67 & 0.38 & 0.86 \\
\textbf{Naive} & 0.51 & 0.86 & 0.73 & 0.69 & 0.39 & 0.67 & 0.38 & 0.71 & 0.34 & 0.87 \\
\midrule

\multicolumn{11}{c}{\textit{Coverage Rate} -- \textbf{CR (\%) $\uparrow$}} \\
\cdashline{1-11}[2.5pt/5pt]\noalign{\vskip 0.5ex}
\textbf{W/o Retrieve} & 0.92 & 0.92 & 0.91 & 0.90 & 0.90 & 0.90 & 0.90 & 0.88 & 0.92 & 0.92 \\
\textbf{FiD} & 0.94 & 0.96 & 0.98 & 0.96 & 0.95 & 0.94 & 0.97 & 0.94 & 0.93 & 0.96 \\
\textbf{Fusion} & 0.90 & 0.90 & 0.89 & 0.89 & 0.90 & 0.89 & 0.92 & 0.89 & 0.91 & 0.93 \\
\textbf{HyDE} & 0.88 & 0.87 & 0.88 & 0.87 & 0.89 & 0.88 & 0.90 & 0.90 & 0.92 & 0.93 \\
\textbf{RAPTOR} & 0.91 & 0.89 & 0.88 & 0.92 & 0.91 & 0.90 & 0.89 & 0.90 & 0.92 & 0.92 \\
\textbf{RAT} & 0.91 & 0.92 & 0.93 & 0.88 & 0.92 & 0.89 & 0.89 & 0.88 & 0.92 & 0.92 \\
\textbf{REPLUG} & 0.93 & 0.97 & 0.97 & 0.90 & 0.88 & 0.90 & 0.92 & 0.95 & 0.96 & 0.98 \\
\textbf{Self} & 0.90 & 0.92 & 0.88 & 0.89 & 0.89 & 0.89 & 0.89 & 0.89 & 0.91 & 0.92 \\
\textbf{Naive} & 0.91 & 0.89 & 0.90 & 0.90 & 0.92 & 0.89 & 0.90 & 0.89 & 0.92 & 0.92 \\
\midrule

\multicolumn{11}{c}{\textit{Prediction Uncertainty} -- \textbf{SS $\downarrow$}} \\
\cdashline{1-11}[2.5pt/5pt]\noalign{\vskip 0.5ex}
\textbf{W/o Retrieve} & 2.72 & 1.65 & 4.68 & 2.67 & 3.76 & 2.98 & 3.48 & 2.59 & 2.55 & 1.52 \\
\textbf{FiD} & 2.85 & 3.77 & 4.82 & 2.85 & 3.98 & 3.69 & 3.85 & 2.83 & 2.84 & 3.77 \\
\textbf{Fusion} & 2.62 & 1.51 & 3.77 & 2.40 & 3.65 & 2.90 & 3.40 & 2.35 & 2.59 & 1.66 \\
\textbf{HyDE} & 2.34 & 1.64 & 4.30 & 2.39 & 3.73 & 3.08 & 3.58 & 2.44 & 2.63 & 1.67 \\
\textbf{RAPTOR} & 2.68 & 1.51 & 3.84 & 2.43 & 3.70 & 2.98 & 3.24 & 2.44 & 2.69 & 1.55 \\
\textbf{RAT} & 2.79 & 2.07 & 4.57 & 2.62 & 3.74 & 2.85 & 3.41 & 2.55 & 2.71 & 1.84 \\
\textbf{REPLUG} & 2.24 & 3.50 & 4.61 & 2.57 & 3.83 & 3.66 & 3.73 & 2.69 & 2.80 & 3.50 \\
\textbf{Self} & 2.50 & 1.56 & 4.07 & 2.48 & 3.67 & 2.95 & 3.35 & 2.50 & 2.63 & 1.59 \\
\textbf{Naive} & 2.64 & 1.48 & 3.79 & 2.34 & 3.74 & 2.91 & 3.21 & 2.40 & 2.69 & 1.55 \\
    \bottomrule
  \end{tabularx}
  \end{adjustbox}
\end{table*}

\Table~\ref{tab:rag_unc_llm_8b_wrong_aware} reports accuracy, coverage, and uncertainty under the Wrong-Aware Prompting condition, where prompts intentionally include misleading confidence information. This experiment evaluates RAG robustness when prompt-level signals conflict with model beliefs or retrieved evidence. The table spans all domains and irrelevant-context test sets. Values quantify how robust each RAG method is when being misled by auxiliary signals.

\begin{table*}[t]
  \centering
  \footnotesize
  \setlength{\tabcolsep}{4pt}
  \caption{\textbf{Uncertainty (RAG) 8B LLM results across tasks through wrong-aware prompting.}}
  \label{tab:rag_unc_llm_8b_wrong_aware}
  \begin{adjustbox}{center, max width=\textwidth}
  \begin{tabularx}{\textwidth}{l *{10}{c}}
    \toprule
    \multicolumn{1}{l}{} & 
    \multicolumn{1}{c}{\textbf{Healthcare}} & 
    \multicolumn{2}{c}{\textbf{Code}} & 
    \multicolumn{1}{c}{\textbf{Research}} & 
    \multicolumn{1}{c}{\textbf{Math}} &
    \multicolumn{3}{c}{\textbf{General Text}} &
    \multicolumn{2}{c}{\textbf{Irrelevant Contexts}} \\
    \cmidrule(lr){2-2}\cmidrule(lr){3-4}\cmidrule(lr){5-5}\cmidrule(lr){6-6}\cmidrule(lr){7-9}\cmidrule(lr){10-11}
    \textbf{RAG} &
    \textbf{Healthver} &
    \textbf{Odex} &
    \textbf{LCA} &
    \textbf{SciFact} &
    \textbf{Olympiad} &
    \textbf{CRAG} &
    \textbf{NewsSum} &
    \textbf{DialFact} &
    \textbf{W/DialFact} &
    \textbf{W/Odex} \\
    \midrule

    \multicolumn{11}{c}{\textit{Performance} -- \textbf{Acc (\%) $\uparrow$}} \\
\cdashline{1-11}[2.5pt/5pt]\noalign{\vskip 0.5ex}
\textbf{W/o Retrieve} & 0.45 & 0.85 & 0.20 & 0.42 & 0.29 & 0.52 & 0.40 & 0.44 & 0.41 & 0.85 \\
\textbf{Fid} & 0.38 & 0.27 & 0.21 & 0.43 & 0.30 & 0.31 & 0.23 & 0.34 & 0.33 & 0.26 \\
\textbf{Fusion} & 0.49 & 0.82 & 0.67 & 0.71 & 0.26 & 0.62 & 0.40 & 0.63 & 0.30 & 0.84 \\
\textbf{HyDE} & 0.50 & 0.79 & 0.56 & 0.68 & 0.38 & 0.57 & 0.40 & 0.67 & 0.32 & 0.79 \\
\textbf{RAPTOR} & 0.48 & 0.83 & 0.67 & 0.70 & 0.29 & 0.61 & 0.38 & 0.68 & 0.34 & 0.85 \\
\textbf{RAT} & 0.50 & 0.63 & 0.27 & 0.44 & 0.33 & 0.53 & 0.36 & 0.52 & 0.36 & 0.62 \\
\textbf{REPLUG} & 0.47 & 0.84 & 0.67 & 0.71 & 0.28 & 0.61 & 0.39 & 0.67 & 0.33 & 0.85 \\
\textbf{Self} & 0.49 & 0.83 & 0.70 & 0.70 & 0.34 & 0.60 & 0.39 & 0.64 & 0.34 & 0.84 \\
\textbf{Naive} & 0.48 & 0.83 & 0.67 & 0.70 & 0.29 & 0.61 & 0.38 & 0.68 & 0.34 & 0.84 \\
\midrule

\multicolumn{11}{c}{\textit{Coverage Rate} -- \textbf{CR (\%) $\uparrow$}} \\
\cdashline{1-11}[2.5pt/5pt]\noalign{\vskip 0.5ex}
\textbf{W/o Retrieve} & 0.72 & 0.88 & 0.60 & 0.71 & 0.73 & 0.84 & 0.70 & 0.73 & 0.73 & 0.88 \\
\textbf{Fid} & 0.95 & 0.96 & 0.95 & 0.98 & 0.95 & 0.95 & 0.97 & 0.95 & 0.93 & 0.96 \\
\textbf{Fusion} & 0.92 & 0.93 & 0.95 & 0.86 & 0.90 & 0.92 & 0.90 & 0.90 & 0.82 & 0.93 \\
\textbf{HyDE} & 0.91 & 0.91 & 0.91 & 0.86 & 0.90 & 0.91 & 0.91 & 0.90 & 0.78 & 0.93 \\
\textbf{RAPTOR} & 0.91 & 0.92 & 0.93 & 0.90 & 0.89 & 0.92 & 0.91 & 0.91 & 0.86 & 0.92 \\
\textbf{RAT} & 0.90 & 0.93 & 0.93 & 0.92 & 0.90 & 0.91 & 0.91 & 0.90 & 0.87 & 0.90 \\
\textbf{REPLUG} & 0.91 & 0.95 & 0.93 & 0.94 & 0.92 & 0.92 & 0.93 & 0.94 & 0.95 & 0.96 \\
\textbf{Self} & 0.90 & 0.93 & 0.93 & 0.92 & 0.90 & 0.91 & 0.91 & 0.90 & 0.87 & 0.93 \\
\textbf{Naive} & 0.91 & 0.92 & 0.93 & 0.90 & 0.90 & 0.92 & 0.91 & 0.91 & 0.86 & 0.92 \\
\midrule

\multicolumn{11}{c}{\textit{Prediction Uncertainty} -- \textbf{SS $\downarrow$}} \\
\cdashline{1-11}[2.5pt/5pt]\noalign{\vskip 0.5ex}
\textbf{W/o Retrieve} & 2.01 & 1.52 & 3.04 & 2.02 & 2.94 & 2.87 & 2.48 & 2.02 & 2.05 & 1.52 \\
\textbf{Fid} & 2.85 & 3.77 & 4.73 & 2.93 & 3.98 & 3.71 & 3.85 & 2.85 & 2.84 & 3.77 \\
\textbf{Fusion} & 2.69 & 1.78 & 2.97 & 1.94 & 3.64 & 2.44 & 2.75 & 2.05 & 2.24 & 1.73 \\
\textbf{HyDE} & 2.54 & 1.79 & 3.05 & 1.94 & 3.51 & 2.59 & 2.90 & 2.02 & 2.24 & 1.92 \\
\textbf{RAPTOR} & 2.66 & 1.77 & 3.12 & 2.09 & 3.60 & 2.44 & 2.83 & 2.14 & 2.34 & 1.76 \\
\textbf{RAT} & 2.63 & 1.89 & 4.65 & 2.60 & 3.57 & 2.76 & 3.34 & 2.41 & 2.52 & 1.83 \\
\textbf{REPLUG} & 2.09 & 3.24 & 4.49 & 2.66 & 3.93 & 3.53 & 3.68 & 2.61 & 2.71 & 3.25 \\
\textbf{Self} & 2.59 & 1.77 & 3.20 & 2.26 & 3.49 & 2.57 & 2.81 & 2.13 & 1.99 & 1.81 \\
\textbf{Naive} & 2.67 & 1.76 & 3.12 & 2.09 & 3.65 & 2.46 & 2.83 & 2.14 & 2.34 & 1.76 \\
    \bottomrule
  \end{tabularx}
  \end{adjustbox}
\end{table*}

\Table~\ref{tab:rag_correct_incorrect_8b} extends the misleading-confidence analysis by jointly considering retrieval-induced noise and perturbed confidence cues. It includes accuracy, coverage, and uncertainty across the full suite of tasks, revealing interaction effects between retrieval irrelevance and deceptive confidence exposure. This table emphasizes how some RAG methods degrade sharply when both retrieval noise and prompt-level misdirection are present.

\begin{table*}[t]
\centering
\footnotesize
\setlength{\tabcolsep}{4pt}
\caption{\textbf{RAG Performance on 8B Model for LLM-Correct vs LLM-Incorrect Cases}.
\cmark: LLM originally correct, \xmark: LLM originally incorrect.}
\label{tab:rag_correct_incorrect_8b}
\begin{adjustbox}{center, max width=\textwidth}
\begin{tabularx}{\textwidth}{c l *{8}{c}}
\toprule
& & \textbf{Healthcare} & \multicolumn{2}{c}{\textbf{Code}} & \textbf{Research} & \textbf{Math} & \multicolumn{3}{c}{\textbf{General Text}} \\
\cmidrule(lr){3-3} \cmidrule(lr){4-5} \cmidrule(lr){6-6} \cmidrule(lr){7-7} \cmidrule(lr){8-10}
\textbf{LLM} & \textbf{RAG} &
\textbf{Healthver} &
\textbf{Odex} &
\textbf{LCA} &
\textbf{SciFact} &
\textbf{Olympiad} &
\textbf{CRAG} &
\textbf{NewsSum} &
\textbf{DialFact} \\
\midrule
\multicolumn{10}{c}{\textit{Performance} — \textbf{Accuracy (\%) $\uparrow$}} \\
\cdashline{1-10}[2pt/4pt]
\cmark & \textbf{Fusion} & 0.82 & 0.96 & 0.88 & 0.90 & 0.63 & 0.80 & 0.65 & 0.86 \\
\xmark & \textbf{Fusion} & 0.28 & 0.18 & 0.80 & 0.52 & 0.24 & 0.50 & 0.25 & 0.59 \\
\cmark & \textbf{HyDE} & 0.83 & 0.96 & 0.71 & 0.82 & 0.56 & 0.78 & 0.71 & 0.84 \\
\xmark & \textbf{HyDE} & 0.28 & 0.11 & 0.74 & 0.60 & 0.31 & 0.42 & 0.27 & 0.61 \\
\cmark & \textbf{RAPTOR} & 0.87 & 0.96 & 0.76 & 0.84 & 0.70 & 0.81 & 0.66 & 0.87 \\
\xmark & \textbf{RAPTOR} & 0.23 & 0.14 & 0.72 & 0.58 & 0.36 & 0.49 & 0.21 & 0.58 \\
\cmark & \textbf{RAT} & 0.79 & 0.93 & 0.88 & 0.86 & 0.62 & 0.82 & 0.67 & 0.76 \\
\xmark & \textbf{RAT} & 0.25 & 0.21 & 0.20 & 0.48 & 0.37 & 0.50 & 0.26 & 0.54 \\
\cmark & \textbf{REPLUG} & 0.89 & 0.96 & 0.76 & 0.86 & 0.61 & 0.82 & 0.66 & 0.87 \\
\xmark & \textbf{REPLUG} & 0.21 & 0.14 & 0.72 & 0.58 & 0.23 & 0.49 & 0.21 & 0.56 \\
\cmark & \textbf{Self-RAG} & 0.82 & 0.94 & 0.82 & 0.87 & 0.79 & 0.81 & 0.71 & 0.84 \\
\xmark & \textbf{Self-RAG} & 0.26 & 0.11 & 0.72 & 0.57 & 0.20 & 0.42 & 0.18 & 0.54 \\
\cmark & \textbf{Naive} & 0.88 & 0.96 & 0.71 & 0.84 & 0.73 & 0.82 & 0.65 & 0.87 \\
\xmark & \textbf{Naive} & 0.27 & 0.14 & 0.77 & 0.58 & 0.23 & 0.50 & 0.22 & 0.59 \\
\midrule
\multicolumn{10}{c}{\textit{Uncertainty} — \textbf{Set Size (SS) $\downarrow$}} \\
\cdashline{1-10}[2pt/4pt]
\cmark & \textbf{Fusion} & 2.60 & 1.69 & 2.26 & 1.96 & 3.35 & 2.18 & 2.79 & 1.90 \\
\xmark & \textbf{Fusion} & 2.67 & 2.02 & 2.17 & 2.36 & 3.36 & 2.62 & 2.84 & 2.08 \\
\cmark & \textbf{HyDE} & 2.38 & 1.63 & 2.56 & 1.86 & 3.68 & 2.24 & 2.72 & 1.88 \\
\xmark & \textbf{HyDE} & 2.57 & 2.00 & 2.33 & 2.12 & 3.69 & 2.63 & 2.74 & 2.04 \\
\cmark & \textbf{RAPTOR} & 2.56 & 1.68 & 2.82 & 1.77 & 3.60 & 2.11 & 2.60 & 1.96 \\
\xmark & \textbf{RAPTOR} & 2.72 & 1.95 & 2.67 & 2.14 & 3.72 & 2.57 & 2.74 & 2.12 \\
\cmark & \textbf{RAT} & 2.58 & 1.67 & 4.59 & 2.45 & 3.17 & 2.32 & 3.01 & 2.19 \\
\xmark & \textbf{RAT} & 2.58 & 1.91 & 4.44 & 2.48 & 3.37 & 2.73 & 3.06 & 2.25 \\
\cmark & \textbf{REPLUG} & 2.34 & 3.50 & 4.32 & 2.52 & 3.77 & 3.69 & 3.71 & 2.73 \\
\xmark & \textbf{REPLUG} & 2.16 & 3.48 & 4.71 & 2.61 & 3.89 & 3.76 & 3.75 & 2.65 \\
\cmark & \textbf{Self-RAG} & 2.66 & 1.74 & 3.00 & 2.08 & 3.44 & 2.31 & 2.61 & 1.97 \\
\xmark & \textbf{Self-RAG} & 2.71 & 1.95 & 2.48 & 2.50 & 3.54 & 2.77 & 2.69 & 2.12 \\
\cmark & \textbf{Naive} & 2.59 & 1.66 & 2.59 & 1.75 & 3.50 & 2.08 & 2.71 & 1.97 \\
\xmark & \textbf{Naive} & 2.69 & 1.89 & 2.42 & 2.08 & 3.56 & 2.55 & 2.80 & 2.13 \\
\bottomrule
\end{tabularx}
\end{adjustbox}
\end{table*}

\Table~\ref{tab:rag_unc_llm_3b} provides accuracy, coverage, and uncertainty for all RAG methods using a 3B-parameter LLM, serving as a smaller-model counterpart to the main 8B results. It covers all domains and includes irrelevant-context evaluations. Subscripts indicate uncertainty calibrated by the LLM, enabling direct comparison of how retrieval affects uncertainty differently at a smaller scale. This table highlights model-size effects on RAG robustness, calibration, and sensitivity to retrieval noise.

\begin{table*}[t]
  \centering
  \footnotesize
  \setlength{\tabcolsep}{4pt}
  \caption{\textbf{Uncertainty (RAG) 3B LLM results across tasks through normal prompting.}}
  \label{tab:rag_unc_llm_3b}
  \begin{adjustbox}{center, max width=\textwidth}
  \begin{tabularx}{\textwidth}{l *{10}{c}}
    \toprule
    \multicolumn{1}{l}{} & 
    \multicolumn{1}{c}{\textbf{Healthcare}} & 
    \multicolumn{2}{c}{\textbf{Code}} & 
    \multicolumn{1}{c}{\textbf{Research}} & 
    \multicolumn{1}{c}{\textbf{Math}} &
    \multicolumn{3}{c}{\textbf{General Text}} &
    \multicolumn{2}{c}{\textbf{Irrelevant Contexts}} \\
    \cmidrule(lr){2-2}\cmidrule(lr){3-4}\cmidrule(lr){5-5}\cmidrule(lr){6-6}\cmidrule(lr){7-9}\cmidrule(lr){10-11}
    \textbf{RAG} &
    \textbf{Healthver} &
    \textbf{Odex} &
    \textbf{LCA} &
    \textbf{SciFact} &
    \textbf{Olympiad} &
    \textbf{CRAG} &
    \textbf{NewsSum} &
    \textbf{DialFact} &
    \textbf{W/DialFact} &
    \textbf{W/Odex} \\
    \midrule

    \multicolumn{11}{c}{\textit{Performance} -- \textbf{Acc (\%) $\uparrow$}} \\
\cdashline{1-11}[2.5pt/5pt]\noalign{\vskip 0.5ex}
\textbf{W/o Retrieve} & 0.46 & 0.82 & 0.18 & 0.45 & 0.29 & 0.52 & 0.37 & 0.31 & 0.27 & 0.81 \\
\textbf{Fusion} & 0.43 & 0.84 & 0.63 & 0.63 & 0.32 & 0.59 & 0.37 & 0.49 & 0.24 & 0.77 \\
\textbf{HyDE} & 0.53 & 0.81 & 0.61 & 0.47 & 0.30 & 0.55 & 0.41 & 0.58 & 0.28 & 0.79 \\
\textbf{RAPTOR} & 0.41 & 0.82 & 0.63 & 0.60 & 0.28 & 0.58 & 0.39 & 0.46 & 0.24 & 0.80 \\
\textbf{RAT} & 0.37 & 0.79 & 0.43 & 0.56 & 0.35 & 0.61 & 0.32 & 0.48 & 0.33 & 0.80 \\
\textbf{REPLUG} & 0.48 & 0.83 & 0.68 & 0.61 & 0.28 & 0.59 & 0.39 & 0.45 & 0.26 & 0.82 \\
\textbf{Self} & 0.47 & 0.80 & 0.63 & 0.58 & 0.29 & 0.56 & 0.41 & 0.48 & 0.28 & 0.77 \\
\textbf{Naive} & 0.41 & 0.82 & 0.63 & 0.60 & 0.28 & 0.59 & 0.39 & 0.46 & 0.24 & 0.80 \\
\midrule

\multicolumn{11}{c}{\textit{Coverage Rate} -- \textbf{CR (\%) $\uparrow$}} \\
\cdashline{1-11}[2.5pt/5pt]\noalign{\vskip 0.5ex}
\textbf{W/o Retrieve} & 0.92 & 0.94 & 0.85 & 0.92 & 0.91 & 0.91 & 0.88 & 0.89 & 0.94 & 0.94 \\
\textbf{Fusion} & 0.92 & 0.94 & 0.95 & 0.87 & 0.93 & 0.91 & 0.91 & 0.88 & 0.90 & 0.94 \\
\textbf{HyDE} & 0.94 & 0.92 & 0.91 & 0.89 & 0.91 & 0.89 & 0.90 & 0.90 & 0.95 & 0.92 \\
\textbf{RAPTOR} & 0.91 & 0.91 & 0.93 & 0.90 & 0.91 & 0.91 & 0.90 & 0.89 & 0.92 & 0.91 \\
\textbf{RAT} & 0.92 & 0.93 & 0.87 & 0.91 & 0.87 & 0.90 & 0.89 & 0.89 & 0.89 & 0.93 \\
\textbf{REPLUG} & 0.93 & 0.98 & 0.90 & 0.97 & 0.88 & 0.95 & 0.97 & 0.97 & 0.98 & 0.97 \\
\textbf{Self} & 0.90 & 0.92 & 0.91 & 0.89 & 0.89 & 0.90 & 0.90 & 0.88 & 0.90 & 0.90 \\
\textbf{Naive} & 0.93 & 0.92 & 0.94 & 0.91 & 0.91 & 0.91 & 0.90 & 0.89 & 0.92 & 0.91 \\
\midrule

\multicolumn{11}{c}{\textit{Prediction Uncertainty} -- \textbf{SS $\downarrow$}} \\
\cdashline{1-11}[2.5pt/5pt]\noalign{\vskip 0.5ex}
\textbf{W/o Retrieve} & 2.74 & 1.68 & 4.17 & 2.74 & 3.79 & 2.95 & 3.34 & 2.60 & 2.73 & 1.67 \\
\textbf{Fusion} & 2.69 & 1.72 & 3.15 & 2.31 & 3.83 & 2.61 & 3.15 & 2.20 & 2.67 & 1.71 \\
\textbf{HyDE} & 2.53 & 1.71 & 3.42 & 2.39 & 3.78 & 2.79 & 3.23 & 2.13 & 2.70 & 1.71 \\
\textbf{RAPTOR} & 2.59 & 1.64 & 3.28 & 2.39 & 3.75 & 2.64 & 3.12 & 2.28 & 2.63 & 1.64 \\
\textbf{RAT} & 2.80 & 1.73 & 4.23 & 2.48 & 3.58 & 2.73 & 3.29 & 2.51 & 2.63 & 1.73 \\
\textbf{REPLUG} & 2.44 & 3.63 & 4.53 & 2.88 & 3.85 & 3.65 & 3.80 & 2.82 & 2.96 & 3.62 \\
\textbf{Self} & 2.62 & 1.67 & 3.43 & 2.48 & 3.73 & 2.72 & 3.13 & 2.31 & 2.61 & 1.68 \\
\textbf{Naive} & 2.61 & 1.65 & 3.26 & 2.36 & 3.76 & 2.66 & 3.09 & 2.31 & 2.62 & 1.64 \\
    \bottomrule
  \end{tabularx}
  \end{adjustbox}
\end{table*}

\Tableref{tab:rag_results_reorganized} shows the scenario when the retrieval system returns irrelevant documents and correct documents for each query. In this experiment, we retrieve ten relevant documents and 10 irrelevant documents from the retrieval system for each query, which significantly affects the performance and uncertainty of RAG systems.

\begin{table*}[t]
  \centering
  \footnotesize
  \setlength{\tabcolsep}{3.5pt}
  \caption{\textbf{Accuracy, Coverage, and Uncertainty results of different RAG methods on three datasets.} The retrieval system is poisoned and returns irrelevant documents. Subscripts show differences from normal context (irrelevant - normal).}
  \label{tab:rag_results_reorganized}
  \begin{adjustbox}{center, max width=0.95\textwidth}
  \begin{tabular}{lccccccccc}
    \toprule
    & \multicolumn{9}{c}{\textbf{Model}} \\
    \cmidrule(l){2-10}
    \textbf{Dataset} & 
    \rotatebox[origin=c]{0}{\textbf{W/o}} & 
    \textbf{Fid} & 
    \textbf{Fusion} & 
    \textbf{HyDE} & 
    \textbf{RAPTOR} & 
    \textbf{RAT} & 
    \textbf{REPLUG} & 
    \textbf{Self} & 
    \textbf{Naive} \\
    \cmidrule(lr){1-1}\cmidrule(l){2-10}
    \textbf{Odex} \\
    \quad Acc (\%) $\uparrow$ & 0.88$_{+0.00}$ & 0.27$_{-0.01}$ & 0.84$_{-0.02}$ & 0.86$_{+0.01}$ & 0.85$_{+0.00}$ & 0.85$_{+0.01}$ & 0.85$_{-0.01}$ & 0.85$_{+0.02}$ & 0.85$_{-0.01}$ \\
    \quad CR (\%) $\uparrow$   & 0.92$_{+0.00}$ & 0.94$_{-0.01}$ & 0.93$_{-0.01}$ & 0.92$_{-0.01}$ & 0.93$_{+0.00}$ & 0.92$_{+0.00}$ & 0.97$_{+0.00}$ & 0.93$_{+0.01}$ & 0.93$_{+0.01}$ \\
    \quad SS $\downarrow$     & 1.66$_{+0.00}$ & 3.77$_{+0.14}$ & 1.67$_{-0.06}$ & 1.70$_{+0.02}$ & 1.64$_{-0.07}$ & 1.69$_{-0.01}$ & 3.50$_{+0.00}$ & 1.72$_{-0.05}$ & 1.64$_{-0.05}$ \\\hline
    \addlinespace[0.3em]
    \textbf{LCA} \\
    \quad Acc (\%) $\uparrow$ & 0.21$_{+0.00}$ & 0.18$_{-0.03}$ & 0.22$_{-0.60}$ & 0.29$_{-0.44}$ & 0.24$_{-0.49}$ & 0.18$_{-0.16}$ & 0.23$_{-0.50}$ & 0.20$_{-0.54}$ & 0.22$_{-0.54}$ \\
    \quad CR (\%) $\uparrow$   & 0.91$_{+0.00}$ & 0.97$_{+0.03}$ & 0.91$_{-0.01}$ & 0.84$_{-0.07}$ & 0.92$_{-0.02}$ & 0.95$_{+0.02}$ & 0.97$_{+0.00}$ & 0.93$_{-0.01}$ & 0.92$_{-0.01}$ \\
    \quad SS $\downarrow$     & 4.64$_{+0.00}$ & 4.82$_{+0.06}$ & 4.45$_{+2.26}$ & 3.90$_{+1.52}$ & 4.47$_{+1.77}$ & 4.64$_{+0.17}$ & 4.91$_{+0.28}$ & 4.55$_{+1.96}$ & 4.47$_{+2.01}$ \\\hline
    \addlinespace[0.3em]
    \textbf{DialFact} \\
    \quad Acc (\%) $\uparrow$ & 0.47$_{+0.00}$ & 0.34$_{-0.01}$ & 0.47$_{-0.24}$ & 0.44$_{-0.28}$ & 0.47$_{-0.25}$ & 0.52$_{-0.12}$ & 0.47$_{-0.24}$ & 0.39$_{-0.29}$ & 0.47$_{-0.25}$ \\
    \quad CR (\%) $\uparrow$   & 0.93$_{+0.00}$ & 0.94$_{+0.00}$ & 0.89$_{-0.02}$ & 0.90$_{+0.00}$ & 0.90$_{-0.01}$ & 0.90$_{+0.00}$ & 0.96$_{+0.01}$ & 0.89$_{-0.02}$ & 0.90$_{+0.00}$ \\
    \quad SS $\downarrow$     & 2.55$_{+0.00}$ & 2.83$_{-0.01}$ & 2.33$_{+0.33}$ & 2.42$_{+0.45}$ & 2.36$_{+0.31}$ & 2.49$_{+0.27}$ & 2.75$_{+0.06}$ & 2.48$_{+0.43}$ & 2.36$_{+0.31}$ \\
    \bottomrule
  \end{tabular}
  \end{adjustbox}
\end{table*}

\begin{table*}[t]
  \centering
  \footnotesize
  \setlength{\tabcolsep}{4pt}
  \caption{\textbf{Statistics of datasets used in the benchmark.}}
  \label{tab:database_statistics}
  \begin{tabular}{l r r r r r r r r}
    \toprule
    \textbf{Dataset file} & \textbf{Samples} & \textbf{Total docs} & \textbf{Avg docs/sample} & \textbf{Total text chars} & \textbf{Avg chars/doc} & \textbf{Total words} & \textbf{Avg words/doc} \\
    \midrule
    OlympiadBench & 661 & 20,672 & N/A (shared) & 51,167,805 & 2,475.22 & 8,910,741 & 431.05 \\
    LCA & 163 & 163 & 1.00 & 42,177,274 & 258,756.28 & 3,879,366 & 23,799.79 \\
    CRAG & 2,330 & 11,650 & 5.00 & 4,208,072,919 & 361,207.98 & 179,542,626 & 15,411.38 \\
    DialFact & 2,000 & 2,435 & 1.22 & 11,835,703 & 4,860.66 & 1,877,746 & 771.15 \\
    Healthver & 1,332 & 4,191 & 3.15 & 88,159,278 & 21,035.38 & 13,167,923 & 3,141.95 \\
    NewsSum & 950 & 950 & 1.00 & 6,109,017 & 6,430.54 & 993,322 & 1,045.60 \\
    Odex & 439 & 34,003 & N/A (shared) & 60,254,653 & 1,772.04 & 8,113,623 & 238.61 \\
    SciFact & 374 & 395 & 1.06 & 26,929,250 & 68,175.32 & 3,872,744 & 9,804.42 \\
    W/Odex & 439 & 3,512 (shared) & N/A (shared) & 4,666,469 & 1,328.72 & 818,938 & 233.18 \\
    W/DialFact & 1000 & 3,512 (shared) & N/A (shared) & 4,666,469 & 1,328.72 & 818,938 & 233.18 \\
    \bottomrule
  \end{tabular}
\end{table*}


\begin{figure*}[t]
\begin{promptbox}[Naive prompt for generating a fake answer]\label{prompt:fake_ans_naive}

You will generate a fake answer for a RAG MCQA dataset.\\

\textbf{Given:}
\begin{itemize}
    \item Original question: ``$\left\{\small\texttt{question}\right\}$''
    \item Correct answer: ``$\left\{\small\texttt{correct\_answer}\right\}$''
\end{itemize}

Output must follow this JSON format:

\begin{tcolorbox}[colback=white, colframe=gray!50, boxrule=0.5pt, arc=0mm, left=2pt, right=2pt, top=2pt, bottom=2pt, fontupper=\footnotesize\ttfamily]

$\{$\\
  ``fake\_answer'': ``...''\\
$\}$

\end{tcolorbox}

Do \textbf{NOT} output anything else.
\end{promptbox}
\end{figure*}

\begin{figure*}[t]
\begin{promptbox}[Prompt for generating a fake answer]\label{prompt:fake_ans}
You will generate a fake answer and a fake supporting document for a RAG MCQA dataset.\\

\textbf{Given:}
\begin{itemize}
    \item Original document: ``$\left\{\small\texttt{original\_document}\right\}$''
    \item Original question: ``$\left\{\small\texttt{question}\right\}$''
    \item Correct answer: ``$\left\{\small\texttt{correct\_answer}\right\}$''
\end{itemize}

\textbf{Your tasks:}

\begin{itemize}
    \item Generate \textbf{ONE} fake answer that:
    \begin{itemize}
        \item Is \textbf{NOT} the correct answer.
        \item Is the same semantic type as the correct answer (e.g., person $\rightarrow$ person).
        \item Is plausible given the question.
        \item Is similar to the correct answer in one way: choose \textbf{ONE} similarity dimension: (role-similar, time-similar, lexical-similar, topic-similar).
        \item Does \textbf{NOT} conflict with safety constraints (no defamation, no sensitive claims).
    \end{itemize}
    \item Generate \textbf{ONE} fake supporting document that:
    \begin{itemize}
        \item Looks realistic and authoritative (e.g., newspaper, government report).
        \item Provides plausible context for the fake answer.
        \item Does \textbf{NOT} mention the correct answer.
        \item CONTRADICTS the original document’s claim.
        \item Avoids sensational, harmful, or political misinformation tone.
        \item Has internal coherence (dates, events, roles must be consistent).
    \end{itemize}
    \item Output must follow this JSON format:
\end{itemize}

\begin{tcolorbox}[colback=white, colframe=gray!50, boxrule=0.5pt, arc=0mm, left=2pt, right=2pt, top=2pt, bottom=2pt, fontupper=\footnotesize\ttfamily]

$\{$\\
  ``fake\_document\_title'': ``...'',\\
  ``fake\_document\_excerpt'': ``...'',\\
  ``fake\_answer'': ``...'',\\
  ``similarity\_type'': ``...''\\
$\}$

\end{tcolorbox}

Do \textbf{NOT} output anything else.
\end{promptbox}
\end{figure*}

\begin{figure*}[t]
\begin{promptbox}[Prompt for regenerating a fake answer]\label{prompt:regen_fake_ans}
You will generate a fake answer and a fake supporting document for a RAG MCQA dataset.\\

\textbf{Given:}
\begin{itemize}
    \item Original document: ``$\left\{\small\texttt{original\_document}\right\}$''
    \item Original question: ``$\left\{\small\texttt{question}\right\}$''
    \item Correct answer: ``$\left\{\small\texttt{correct\_answer}\right\}$''
\end{itemize}

\textbf{Your tasks:}

\begin{itemize}
    \item Generate \textbf{ONE} fake answer that:
    \begin{itemize}
        \item Is \textbf{NOT} the correct answer.
        \item Is the same semantic type as the correct answer (e.g., person $\rightarrow$ person).
        \item Is plausible given the question.
        \item Is similar to the correct answer in one way: choose \textbf{ONE} similarity dimension: (role-similar, time-similar, lexical-similar, topic-similar).
        \item Does \textbf{NOT} conflict with safety constraints (no defamation, no sensitive claims).
    \end{itemize}
    \item Generate \textbf{ONE} fake supporting document that:
    \begin{itemize}
        \item Looks realistic and authoritative (e.g., newspaper, government report).
        \item Provides plausible context for the fake answer.
        \item Does \textbf{NOT} mention the correct answer.
        \item CONTRADICTS the original document’s claim.
        \item Avoids sensational, harmful, or political misinformation tone.
        \item Has internal coherence (dates, events, roles must be consistent).
    \end{itemize}
    \item Output must follow this JSON format:
\end{itemize}

\begin{tcolorbox}[colback=white, colframe=gray!50, boxrule=0.5pt, arc=0mm, left=2pt, right=2pt, top=2pt, bottom=2pt, fontupper=\footnotesize\ttfamily]

$\{$\\
  ``fake\_answer'': ``...'',\\
  ``similarity\_type'': ``...'',\\
  ``fake\_document\_title'': ``...'',\\
  ``fake\_document\_excerpt'': ``...''\\
$\}$

\end{tcolorbox}

\textcolor{red}{Previously, you generated ``$\left\{\small\texttt{old\_incorrect\_answer}\right\}$'' but it's not good enough. Please generate a more confusing answer.}

Do \textbf{NOT} output anything else.
\end{promptbox}
\end{figure*}

\begin{figure*}[t]
\begin{promptbox}[Prompt for answer MCQs]\label{prompt:mcqa}
You are given a multiple-choice question and a set of retrieved context passages.

Answer the question using only the provided context.

Do not explain your reasoning.

\textbf{Context:}\\
$\{\texttt{context}\}$

\textbf{Question:}\\
$\{\texttt{question}\}$

\textbf{Answer format:}\\
$\{\texttt{Answer|X}\}$
\end{promptbox}
\end{figure*}

\begin{figure*}[t]
\begin{promptbox}[Prompt for answer MCQs in the self-aware and wrong-aware experiments.]\label{prompt:mcqa-wrong-aware}
You are given a multiple-choice question and a set of retrieved context passages.

Answer the question using only the provided context.

Knowing that your previous answer had the following confidence:

A. $\{\texttt{Confidence of option A}\}$

B. $\{\texttt{Confidence of option B}\}$

C. $\{\texttt{Confidence of option C}\}$

D. $\{\texttt{Confidence of option D}\}$

Do not explain your reasoning.

\textbf{Context:}\\
$\{\texttt{context}\}$

\textbf{Question:}\\
$\{\texttt{question}\}$

\textbf{Answer format:}\\
$\{\texttt{Answer|X}\}$
\end{promptbox}
\end{figure*}

\begin{figure*}[t]
\begin{promptbox}[Prompt for generating a hypothetical document]\label{prompt:hyde}

You are a helpful assistant that writes comprehensive and informative passages to answer questions.

\vspace{0.5em}

Write a detailed, factual passage that would answer the following question.

\vspace{0.5em}

\textbf{Question:} \{\texttt{question}\}

\vspace{0.5em}

Provide a comprehensive answer with relevant facts and information.

\end{promptbox}
\end{figure*}

\begin{figure*}[t]
\begin{promptbox}[Independent prompts for diverse query generation in Fusion RAG.]\label{prompt:fusion-rag-query}

\textbf{System prompt (randomly selected):}
\begin{itemize}
    \item You are a helpful assistant that rephrases questions while keeping the same meaning.
    \item You are a helpful assistant that creates alternative question formulations.
    \item You are a helpful assistant that generates related questions.
    \item You are a helpful assistant that makes questions more specific.
    \item You are a helpful assistant that simplifies complex questions.
\end{itemize}

\textbf{User instruction (randomly selected):}
\begin{itemize}
    \item Create an alternative question: $\{\texttt{question}\}$
    \item Generate a related question: $\{\texttt{question}\}$
    \item Make this question more specific: $\{\texttt{question}\}$
    \item Simplify this question: $\{\texttt{question}\}$
    \item What is another way to ask about the same topic as this question: $\{\texttt{question}\}$
    \item Generate a related question that might help answer this question: $\{\texttt{question}\}$
\end{itemize}

\textbf{Output format:}\\
$\{\texttt{Query}\}$
\end{promptbox}
\end{figure*}

\end{document}